\documentclass[10pt,twocolumn,letterpaper]{article}

\usepackage[pagenumbers]{cvpr} 

\usepackage{adjustbox, wrapfig}
\newcommand*{\coin}{\textsc{MegaCoin}}

\usepackage{multirow}

\newcount\Comments  
\usepackage{color}
\newcommand{\kibitz}[2]{\ifnum\Comments=1\textcolor{#1}{#2}\fi}

\definecolor{cvprblue}{rgb}{0.21,0.49,0.74}
\usepackage[pagebackref,breaklinks,colorlinks,allcolors=cvprblue]{hyperref}

\usepackage{makecell}

\usepackage{color}
\usepackage{float}

\def\paperID{9117} 
\def\confName{CVPR}
\def\confYear{2025}

\title{\coin: Enhancing Medium-Grained Color Perception for Vision-Language Models}

\author{Ming-Chang Chiu\\
USC\\
{\tt\small mingchac@usc.edu}
\and
Shicheng Wen\thanks{Work done while at USC.}\\
UC Davis\\
{\tt\small scwen@ucdavis.edu}
\and
Pin-Yu Chen\\
IBM Research\\
{\tt\small pin-yu.chen@ibm.com}
\and
Xuezhe Ma\\
USC\\
{\tt\small xuezhema@usc.edu}
}

\begin{document}
\maketitle
\begin{abstract}
In vision-language models (VLMs), the ability to perceive and interpret color and physical environment is crucial for achieving contextually accurate understanding and interaction. However, despite advances in multimodal modeling, there remains a significant lack of specialized datasets that rigorously evaluate a model's capacity to discern subtle color variations and spatial context---critical elements for situational comprehension and reliable deployment across real-world applications.
Toward that goal, we curate \coin\footnote{Dataset releasted at \href{https://github.com/charismaticchiu/MegaCOIN}{https://github.com/charismaticchiu/MegaCOIN}}, a high-quality, human-labeled dataset based on \emph{real} images with various contextual attributes. \coin~consists of two parts: \coin-Instruct, which serves as a supervised fine-tuning (SFT) dataset for VLMs; and \coin-Bench, an annotated test set that can be used as a stand-alone QA dataset. \coin~provides three annotated features for 220,000 real images: foreground color, background color, and description of an object's physical environment, constituting 660k human annotations. In addition, \coin~can be applied to benchmark domain generalization (DG) algorithms. We explore benchmarking DG methods in the linear probing setup for VLM and show some new insights. Last but not least, we show that VLMs, including GPT-4o, have subpar color recognition capabilities, and fine-tuning with \coin~can result in improved performance on visual evaluation tasks. In certain cases, \coin~fine-tuned small-scale opensource models such as LLaVA and Bunny can outperform closed-source GPT-4o. We hope the utilities of \coin~can shed light on the directions VLMs can improve and provide a more complex platform for domain generalization algorithms. 
\end{abstract}


\begin{figure*}[ht]
    \centering
    \begin{minipage}[b]{0.49\textwidth}
    \includegraphics[width=\linewidth]{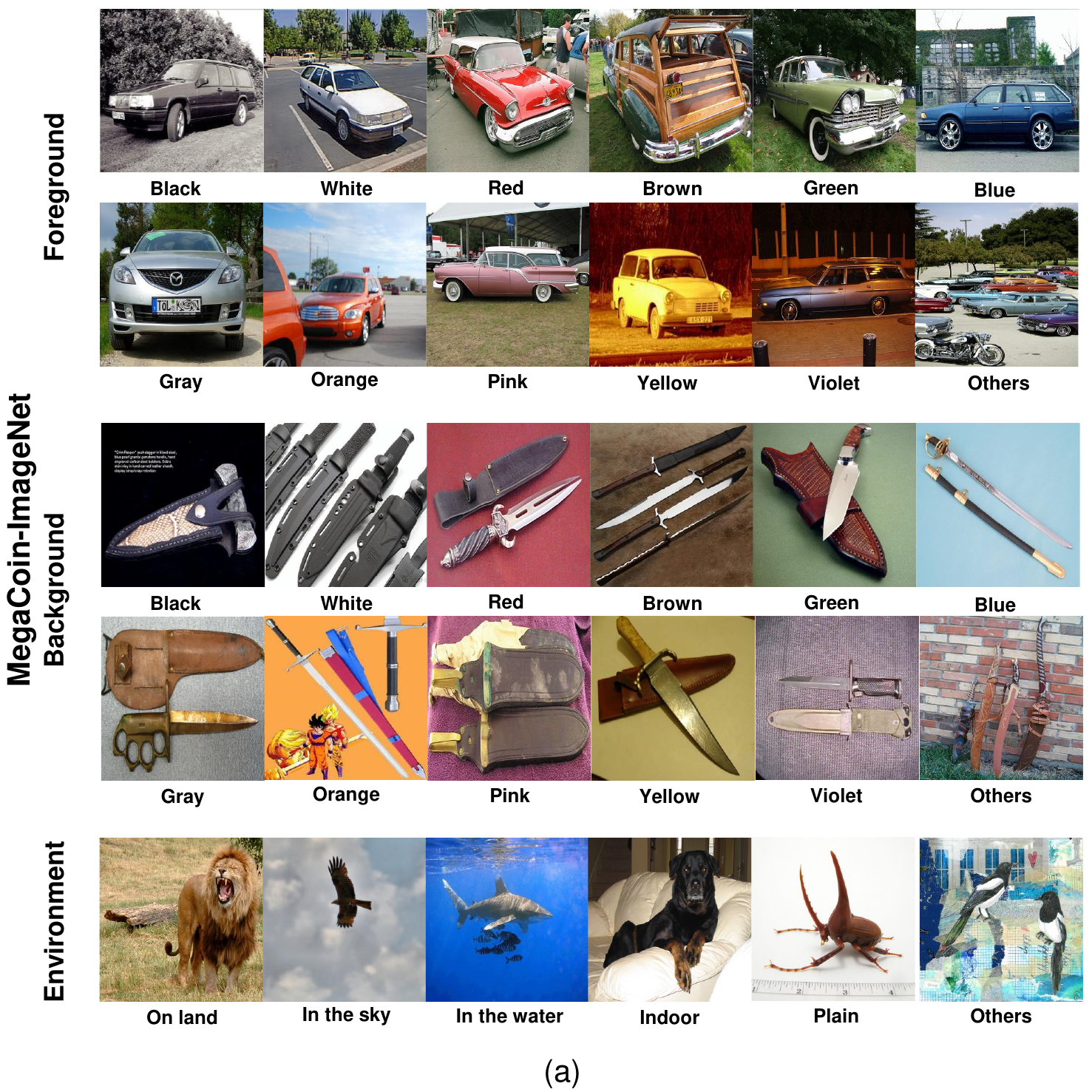}
    \label{fig:megacoin_examples}
    \end{minipage}
    \begin{minipage}[b]{0.25\textwidth}
    \includegraphics[width=\textwidth]{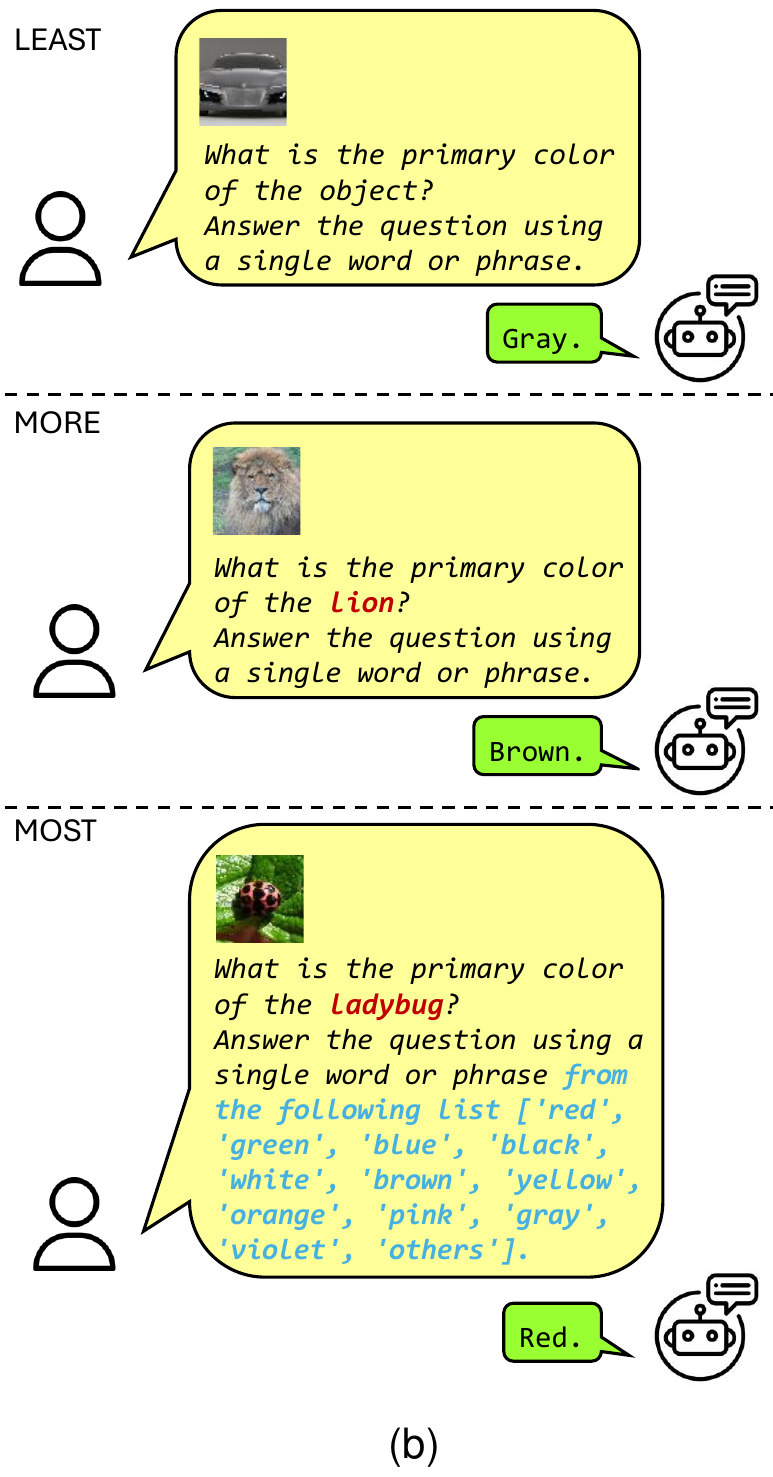}
    \label{fig:benchmark_example}
    \end{minipage}
    \begin{minipage}[b]{0.25\textwidth}

        \includegraphics[width=\textwidth]{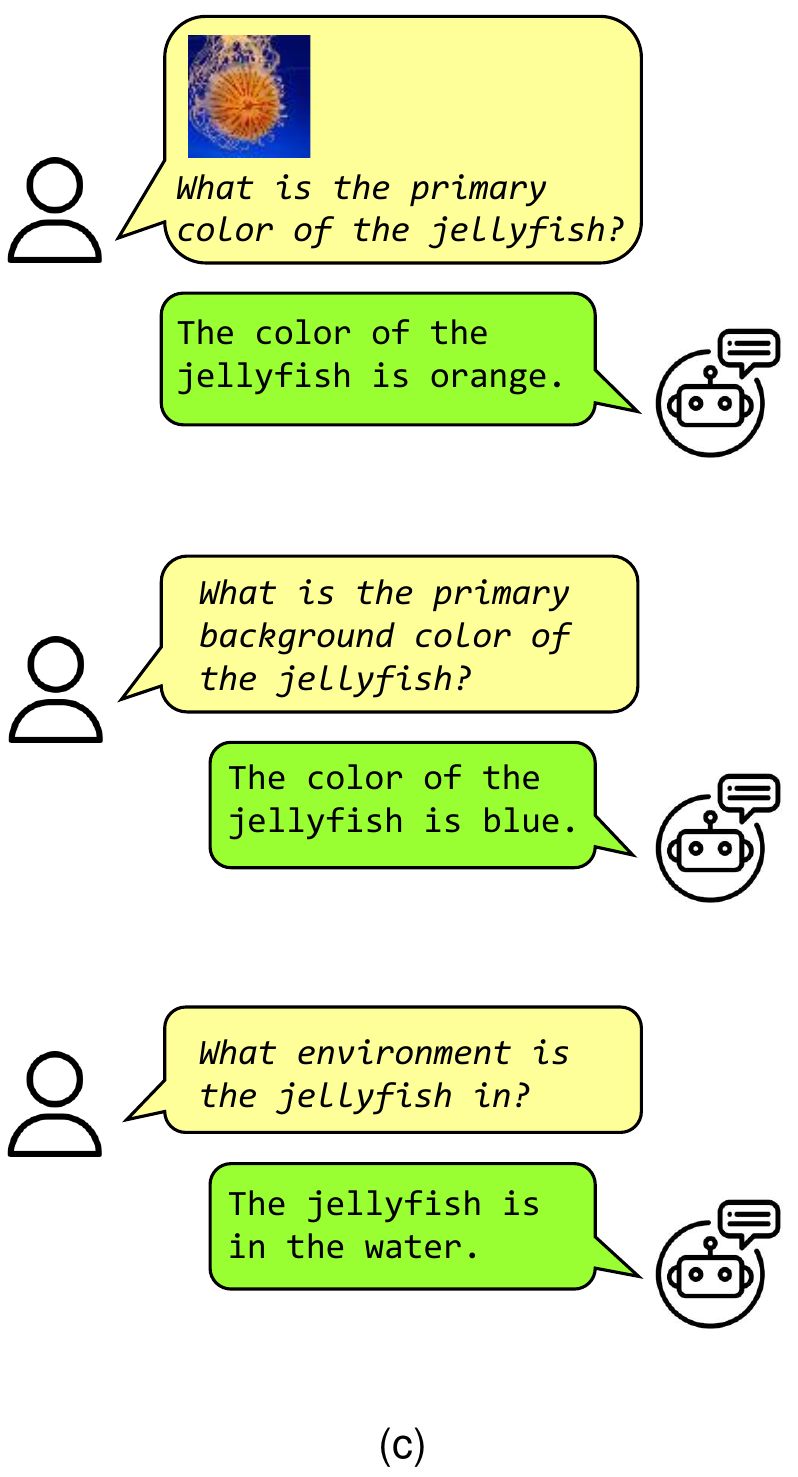}
        \label{fig:instruction_example}
    \end{minipage}\vspace{-5mm}
    \caption{Overview of \coin. (a) Examples of our human-annotated \coin, consisting of three distinct attributes, foreground/background color, physical environment. (b \& c) We use \coin~as an instruction fine-tuning data (\coin-Instruct) and a benchmark (\coin-Bench). (b) Examples of 3-tier \coin-Bench evaluation for a single image. (c) Example of \coin-Instruct SFT pairs for a single image.}\vspace{-5mm}
    \label{fig:overview}
\end{figure*}

\vspace{-2mm}
\section{Introduction}
\vspace{-2mm}
\label{sec:intro}
The rapid advancement of vision-language models (VLMs) has brought about an increasing demand of high-quality datasets that can facilitate improved training and evaluation of these models. A critical factor influencing VLM performance is the quality and granularity of the training data. While existing datasets have substantially contributed to progress in this domain, there remains exploration in datasets that provide medium-grained annotations, which can also capture the complex relationships between visual content and textual descriptions.
In this work, we introduce \coin, a meticulously curated dataset comprised of high-quality, human-annotated images with various contextual attributes. \coin~is dually motivated by certain deficits of recognition capabilities in computer vision models such as color vision \cite{chiu2023better, chiu2022human} and the need for contextual information in VLM training and evaluation, addressing a key limitation in existing datasets that often lack human-labeled contextual elements. By ``medium granularity," we refer to a level of detail that goes beyond simple object labels but stops short of bounding boxes or pixel-level segmentation. This granularity allows for rich and synthesizable descriptions based on human-annotated labels, striking a balance between detail and generalizability.

\coin~leverages \textit{real} images and provides a realistic and challenging setting for training and evaluating VLMs. Our dataset comprises 220,000 real images, each annotated with three human-provided attributes: foreground color (FGD), background color (BGD), and the physical environment (ENV) in which the object is situated. This results in a total of 660,000 annotations. This rich annotation scheme allows for a deeper image understanding that goes beyond simple image labels, enabling more nuanced SFT for VLMs.
\coin~is designed to serve three primary purposes:

\begin{itemize}
    \item \textit{Multimodal Instruction Tuning}: \coin~can serve as an SFT dataset for VLMs. This resource enhances VLMs' ability to understand and generate accurate textual descriptions of visual content, with a particular focus on contextual elements often overlooked in existing datasets. By incorporating medium-grained color and environmental annotations, \coin~enables VLMs to develop a more nuanced understanding of visual scenes.
    \item \textit{Evaluation Benchmark}: The annotated test set of \coin~can be extracted and used as a standalone QA dataset, \coin-Bench. This new benchmark provides a robust tool for evaluating the performance of VLMs on basic visual perception tasks, with a specific emphasis on contextual understanding and color perception.
    \item \textit{Domain Generalization (DG)}: \coin~can be used to explore the effectiveness of DG algorithms in different setups. This utility provides valuable insights from comparing which DG algorithms are useful, a critical capability for real-world applications.

\end{itemize}

Our experimental results show rooms for improvement for VLMs in contextual understanding and shed light on the performance of different domain generalization algorithms, revealing MMD\cite{li2018domain}, CORAL\cite{sun2016deep} and ERM\cite{vapnik1991principles} are on par with each other as measured by different metrics in the linear probing setting. These findings not only demonstrate the utility of \coin~for domain generalization research but also provide practical insights for improving the robustness of VLMs across diverse visual contexts. 

In summary, our contributions in this work include:

\begin{itemize}
    \item \coin-Bench (Sec.~\ref{sec:bench}), a new QA benchmark and a novel Tiered-Multiple Choice QA (Tiered-MQA) scheme for evaluating VLMs' visual perception and contextual understanding capabilities.
    \item \coin-Instruct (Sec.~\ref{sec:instruct}), a novel dataset that enhances visual alignment for VLMs with contextual medium-grained annotations and that can train a 13B or 8B model like LLaVA or Bunny to surpass GPT-4o.
    
    \item New benchmark for DG algorithms (Sec.~\ref{sec:DG}), showcasing \coin's potential to facilitate advancements in this critical area of machine learning research.
\end{itemize}

Through \coin, we aim to bridge the gap between existing datasets and the needs of advanced VLMs, providing a resource that enables more contextually-aware, robust, and generalizable VLMs.


\section{Related Works}\label{sec:relatedworks}


\vspace{-2mm}
\paragraph{Multimodal SFT datasets for VLMs} 
SFT datasets play a crucial role in the development of VLMs. Current approaches to VLM SFT typically involve exposing the model to a diverse set of task-specific instructions and corresponding image-text pairs \cite{liu2023llava, liu2024improved}. Some popular SFT datasets for VLMs include functions like captioning \cite{lin2014microsoft, sharegpt4o}, visual reasoning \cite{liu2023aligning, hudson2019gqa, johnson2017clevr}, chart \cite{liu2023mmc}, science \cite{lu2022learn, kembhavi2016diagram}, OCR \cite{mishra2019ocr, singh2019towards}, etc. This process allows the model to learn to follow natural language instructions and generate appropriate responses based on visual inputs. However, a significant limitation of many existing datasets is their reliance on synthetic data using ChatGPT \cite{liu2024improved, liu2023aligning}. Although synthetic data offer advantages in terms of scalability, they are often costly to construct and lack definitive ground truth. \coin~stands out by offering a dataset of real images with human-annotated attributes. These annotations allow for efficient construction of instruction pairs for SFT while keeping future extensions to produce a guided synthetic version of \coin. 


\vspace{-2mm}
\paragraph{VLM QA datasets} 
Visual Question Answering (VQA) datasets serve as important benchmarks for evaluating the capabilities of VLMs. Several popular test beds have emerged (Tab.~\ref{table:benchmark_comparison}), often spanning multiple domains, such as MME\cite{fu2024mmecomprehensiveevaluationbenchmark}, MMBench \cite{liu2025mmbench}, MM-Vet\cite{yu2023mm}, VQA\cite{antol2015vqa}, GQA\cite{hudson2019gqa}, or more focused capability, such as MMC \cite{liu2023mmc}, MathVista \cite{lu2024mathvistaevaluatingmathematicalreasoning}, etc.

While these benchmarks offer valuable insights into VLM performance across various domains, \coin~takes a different approach by focusing on a fundamental visual capability: color perception and contextual understanding. By providing a dedicated benchmark for color-related queries and environmental context, \coin-Bench aims to serve researchers and practitioners a choice to evaluate basic visual comprehension skills, which has downstream safety implications \cite{furness2003car, chiu2022human}. 


\vspace{-2mm}
\paragraph{Domain generalization dataset}
Domain generalization (DG) is an important area of study in machine learning. Common image benchmarks include ColorMNIST \cite{arjovsky2019invariant}, Spawrious \cite{lynch2023spawrious}, MetaShift \cite{liang2022metashift}, NICO++ \cite{zhang2023nico++}, Waterbirds \cite{sagawa2019distributionally}, CelebA \cite{liu2015deep}, WILDS \cite{beery2020iwildcam}. We summarize and compare some distinctions between these datasets and \coin~in Tab.~\ref{table:DG_datasets_comparison}.

\coin~provides some desiderata of a DG dataset: (1) \textit{multiple} explicitly annotated \textit{Spurious Correlations} (SC), (2) \textit{Real images} that capture natural variations in visual domains, (3) \textit{challenging tasks}: non-binary classification, intra-class heterogeneity and different difficulty levels, (4) can study \textit{Attribute Generalization} (AG) and \textit{Attribute Imbalance} (AI). We refer readers to \cite{qiao2024understanding, yang2023changehardcloserlook, lynch2023spawrious} for detailed definitions.



\begin{table}[t]
\centering
\begin{adjustbox}{width=\linewidth}
\begin{tabular}{lcccc}
\toprule
\textbf{Benchmark} & \textbf{Size} & \textbf{Images} & \textbf{Source} & \textbf{Answer} \\ \midrule
VQA            & $>$1M   & General       & Annotated  & Open     \\ 
GQA            & $>$1M   & General       & Synthesized & Open    \\ 
MME            & 1.5k  & General       & Annotated   & Y/N     \\ 
MMBench        & 3k    & General       & Repurposed  & MQA     \\ 
MM-Vet         & 0.2k  & General       & Repurposed  & MQA     \\ 
MathVista      & 1.4k  & Math          & Synthesized & MQA     \\ 
MMC  & 2k    & Chart/Plot    & Internet, Annotated & Open/MQA \\ 
\coin(ours)       & 210k    & Contextual    & Annotated & Tiered-MQA\\\bottomrule
\end{tabular}
\end{adjustbox}\vspace{-2mm}
\caption{VLM QA Benchmark Comparison.}\vspace{-5mm}
\label{table:benchmark_comparison}
\end{table}

\section{\coin~Dataset}

\begin{table*}
\centering
\footnotesize
\label{tab:color_count}
\begin{adjustbox}{width=\textwidth}
\begin{tabular}{lccccccccccccc}
\toprule
Dataset & split & black & blue & brown & gray & green & orange & others & pink & red & violet & white & yellow \\
\midrule
\multirow[c]{2}{*}{C10FGD} & train & 7414 & 2658 & 12910 & 11085 & 2495 & 896 & 190 & 212 & 2946 & 204 & 7171 & 1819 \\
 & test & 1504 & 518 & 2824 & 1889 & 447 & 176 & 42 & 60 & 547 & 55 & 1638 & 300 \\
\cmidrule{1-14}
\multirow[c]{2}{*}{TIFGD} & train & 17144 & 3960 & 22842 & 21286 & 5715 & 3258 & 5176 & 2175 & 5685 & 931 & 4739 & 7089 \\
& test & 1638 & 387 & 2271 & 1993 & 578 & 315 & 536 & 241 & 605 & 105 & 729 & 602 \\
\cmidrule{1-14}
\multirow[c]{2}{*}{C10BGD} & train & 4830 & 7833 & 6995 & 13129 & 10592 & 454 & 532 & 347 & 642 & 269 & 3408 & 969 \\
 & test & 723 & 1589 & 1460 & 2470 & 2314 & 78 & 227 & 94 & 93 & 60 & 759 & 133 \\
 \cmidrule{1-14}
\multirow[c]{2}{*}{TIBGD} & train & 9172 & 10575 & 15754 & 30844 & 17064 & 909 & 3395 & 1037 & 1267 & 594 & 5864 & 3525 \\
 & test & 894 & 1069 & 1500 & 3079 & 1754 & 83 & 227 & 124 & 124 & 44 & 784 & 318 \\
 \cmidrule{1-14}
 INetFGD & val & 9434 & 2018 & 12371 & 8551 & 2811 & 1477 & 1974 & 863 & 2175 & 337 & 4875 & 3114 \\
INetBGD & val & 4032 & 4557 & 10052 & 12179 & 9488 & 519 & 1813 & 484 & 632 & 210 & 4277 & 1757 \\
\bottomrule
\end{tabular}
\end{adjustbox}\vspace{-2mm}
\caption{Statistics of \coin---Foreground (FGD) and Background (BGD) Color.}\vspace{-2mm}
\end{table*}
\begin{table}
\label{tab:env_count}
\begin{adjustbox}{width=\linewidth}

\begin{tabular}{lccccccc}
\toprule
 Dataset& split & in the sky & in the water & indoor & on land & others & plain \\

\midrule
\multirow[c]{2}{*}{C10ENV} & train & 2962 & 4931 & 2261 & 36642 & 1006 & 2198 \\
 & test & 642 & 816 & 702 & 6886 & 367 & 587\\
\cmidrule{1-8}
\multirow[c]{2}{*}{TIENV} & train & 615 & 6783 & 19310 & 56789 & 9304 & 7199 \\
 & test & 66 & 648 & 2315 & 5522 & 591 & 858 \\
\cmidrule{1-8}
INetENV & val & 495 & 3609 & 14116 & 25663 & 2775 & 3342 \\
\bottomrule
\end{tabular}
\end{adjustbox}\vspace{-2mm}
\caption{Statistics of \coin---Physical Environment (ENV).}\vspace{-5mm}
\end{table}

\subsection{Dataset Construction}
\coin~is a comprehensive dataset that builds upon three widely-used image collections: CIFAR10 (C10), TinyImageNet (TI), and ImageNet (INet). To create \coin, we utilized all images from C10, the entire TI, and the validation set of INet. This combination provides a diverse range of images across various scales and resolutions, ensuring that \coin~covers a broad spectrum of visual scenarios. C10 represents a good baseline for small-scale object recognition tasks and TI offers a step up in complexity from C10. As for INet, it provides high-resolution, real-world examples across 1,000 diverse categories.

To label the colors of foreground and background, we use the eleven basic colors in English, which are \textit{\{White, Black, Grey, Yellow, Red, Blue, Green, Brown, Pink, Orange, Purple/Violet\}}. And for the physical environments, we put the images into \textit{\{In the Sky, In the Water, Indoor, On Land, Others, Plain\}} groups. 

For each image, the annotations involve three attributes:
\begin{itemize}
    \item Foreground Color (FGD): Annotators identify the primary object of the image by the image's class label and select the dominant color from the predefined color set.
    \item Background Color (BGD): Annotators identify the major color in the background of the primary object.
    \item Physical Environment (ENV): Annotators choose from the six physical environments that the object is situated.
\end{itemize}
We show examples of \coin~annotations in Fig.~\ref{fig:overview}(a).

\subsection{Annotation Process}
We collaborate with an industry data curation company, DATUMO, to annotate the foreground, background, and physical environment for each image. 

We enforced a bespoke quality control process to ensure label quality --- Each image is presented to labelers alongside three predefined attribute sets for annotation. Before submission, labelers are prompted to review and confirm their selections. To maintain high standards, any incorrect selection according to company guidelines triggers an automated notification to the labeler. After repeated errors, the labeler’s access is temporarily paused until an administrator reviews and re-enables their account.

\section{\coin-Bench}\label{sec:bench}
Recent advancements in VLMs have unlocked the ability to perform zero-shot tasks that integrate both vision and language, such as Chart \cite{liu2023mmc}, etc. However, we want to take a step back and explore their fundamental abilities of recognizing colors and physical environments. Thus, an evaluation benchmark is essential to assess the performance of various models on these tasks and guide future research and development. Our dataset, \coin-Bench, constructed from the evaluation split of source images, is designed to fill this gap, offering the following key features: Large-scale benchmark to evaluate VLMs' ability to recognize and interpret visual elements --- color, environment, and object --- both individually and in combination, complementing prior benchmarks having a small number of images in the related categories \cite{liu2025mmbench, fu2024mmecomprehensiveevaluationbenchmark}. \coin~leverages the new annotations and the original labels, to provides a novel evaluation methodology for our tasks of interest, which we call Tiered-Multiple Choice QA, a structured format going beyond a single multiple-choice QA pair.

\vspace{-5mm}
\paragraph{Tiered-Multiple Choice QA (Tiered-MQA)}
For each image, we can provide VLMs three levels of information, to test how well VLMs rely on additional information to recognize FGD, BGD and ENV. Examples of the tiers are illustrated in Fig.~\ref{fig:overview}(b).

\begin{itemize}
    \item \textit{Least}: We directly ask the VLM to answer \textit{``What is the primary \{FGD\} of the object?"} By giving the least amount of information, we test the model's ability to identify the object and then perceive the colors or ENV. 
    \item \textit{More}: We provide main object's class label for the VLM, e.g. \textit{``What is the primary \{FGD\} of the \{Class\}?"} Failure to do perform this task would mean that, in addition to incorrect color perception, the VLM may not be able to recognize what target object is. 
    \item \textit{Most}: We provide both the target object's class label and options of attributes to VLM, such as \textit{``What is the primary \{FGD\} of the \{Class\}. Please choose from \{Red, Green, ..., Black\}."} By giving a constrained set of choice and target object, we expect it would be easier for VLMs to comprehend and output correctly.
\end{itemize}

The resulting \coin-Bench consists of 210k images to test each of the three basic perception task, and each annotation can be used for three tiers of evaluations.

We first test \coin-Bench on the state-of-the-art closed-source VLM, GPT-4o, and discover that it may have low capability in our basic vision tasks (Tab.~\ref{tab:coin_bench}), leading us to leverage the training set of source images to construct \coin-Instruct for SFT.

\section{\coin-Instruct}\label{sec:instruct}
We leverage the training set from the source images of \coin~to develop an SFT dataset for VLMs, referred to as \coin-Instruct. More specifically, we use the three attributes we newly annotate along with the original object label to formulate instruction-following pairs as the SFT dataset. 

Our goal is to investigate how \coin-Instruct can enhance model alignment with multimodal tasks, particularly in recognizing and reasoning about colors, objects, and the physical environments in images.

The specific instruction pairs follow structured templates, as exemplified in Fig.~\ref{fig:overview}. For instance, we use instructions such as \textit{``What is the primary foreground color of the \{Object\}? The primary foreground color of the \{Object\} is \{FGD\}."} And similar templates for identifying background colors and physical environment.

This approach enables efficient creations large scale instruction pairs for the models to learn precise associations and improve its understanding of visual concepts within diverse contexts. \vspace{-2mm}

\vspace{-2mm}
\paragraph{Medium Granularity} \coin~employs a medium-grained annotation scheme, striking a balance between detailed description and broad applicability. This choice is justified by several factors:
\begin{itemize}
    \item VLM Compatibility: This level of detail is suited for SFT of vision-language models, providing informed details without overwhelming the model with overly specific information.
    \item Efficiency: This approach allows for a larger number of annotated images within the given budget, increasing the dataset's overall utility.
    \item Flexibility: medium-grained annotations can be easily aggregated for coarser-grained tasks or used as a basis for more fine-grained analysis if needed (e.g. Sec.~\ref{sec:DG}).
\end{itemize}

The resulting SFT dataset consists of 450k instruction-image pairs, and we summarize \coin's statistics in Tab.~\ref{tab:megacoin_statistics}. This rich annotation scheme provides a solid foundation for basic vision-language tasks, particularly those involving color perception and contextual understanding.

\begin{table}[t]
\centering
\begin{adjustbox}{width=0.5\linewidth}
\begin{tabular}{lr}
\toprule
\textbf{Statistics} & \textbf{Num} \\ \midrule
\coin-Instruct & 450k \\ 
– background & 150k \\ 
– foreground & 150k \\ 
– environment & 150k \\ 
\coin-Bench & 210k \\ 
– background & 70k \\ 
– foreground & 70k \\ 
– environment & 70k \\  \bottomrule
\end{tabular}
\end{adjustbox}\vspace{-2mm}
\caption{\coin~Statistics.}
\label{tab:megacoin_statistics}\vspace{-2mm}
\end{table}

\section{Experiment}

\subsection{Baselines}
We leverage LLaVA-v1.5-13B \cite{liu2024improved} and Bunny-Llama3-v1.1-8B \cite{he2024efficient} as our open-source backbones, and  GPT-4o \cite{achiam2023gpt} for closed model evaluation. LLaVA-1.5 excels in visual reasoning and image captioning by integrating a LLaMA-based language model with vision encoders, making it effective for tasks requiring deep understanding of visual data. Bunny-1.1, designed for efficient multimodal learning, uses high-quality training data to deliver strong visual understanding while minimizing computational costs, making it a versatile choice for VLM tasks. GPT-4o is one of the best multi-modal LLMs that lead performances across various tasks.
\subsection{Experiment Setups}\label{sec:exp_setup}

To fine-tune the open-source models within resource constraint, we employ the Low-Rank Adaptation (LoRA) \cite{hu2021lora} technique. We combine the original fine-tuning dataset of each backbone model with our \coin-Instruct. In addition, based on \coin's source image, we can explore three combinations of \coin-Instruct: CIFAR10 (C10), TinyImageNet (TI), and the combination of CIFAR10 and TinyImageNet (C10+TI). For each backbone model, we evaluate four variants based on the SFT data: the original backbone models as baselines, and the three using the aforementioned data combinations. This approach enables us to assess the impact of our dataset on model performance systematically. All models were fine-tuned using 8 NVIDIA A40 GPUs.


\begin{table*}[htbp]
    \centering
    \begin{adjustbox}{width=\textwidth}
\begin{tabular}{cllccccccccc}
\toprule
\multicolumn{2}{c}{Model} & Tier & C10FGD & C10BGD & C10ENV & TIFGD & TIBGD & TIENV & INetFGD & INetBGD & INetENV \\
 \midrule
 \multirow{3}{*}{\rotatebox{90}{Closed}} & \multirow{3}{*}{\textbf{GPT-4o}}
&LEAST & 61.6 {\scriptsize\color{blue}(+0.0)} & 58.29 {\scriptsize\color{blue}(+0.0)} & 30.19 {\scriptsize\color{blue}(+0.0)} & 49.32 {\scriptsize\color{blue}(+0.0)}& 53.45 {\scriptsize\color{blue}(+0.0)}& 26.08 {\scriptsize\color{blue}(+0.0)}& 52.87 {\scriptsize\color{blue}(+0.0)}& 53.75 {\scriptsize\color{blue}(+0.0)}& 29.85 {\scriptsize\color{blue}(+0.0)}\\
& &MORE & 60.04 {\scriptsize\color{blue}(+0.0)}& 58.24 {\scriptsize\color{blue}(+0.0)}& 30.74 {\scriptsize\color{blue}(+0.0)}& 42.79 {\scriptsize\color{blue}(+0.0)}& 49.89 {\scriptsize\color{blue}(+0.0)}& 26.71 {\scriptsize\color{blue}(+0.0)}& 46.08 {\scriptsize\color{blue}(+0.0)}& 50.96 {\scriptsize\color{blue}(+0.0)}& 31.37 {\scriptsize\color{blue}(+0.0)}\\
& &MOST & 65.37 {\scriptsize\color{blue}(+0.0)}& 62.44 {\scriptsize\color{blue}(+0.0)}& 76.2 {\scriptsize\color{blue}(+0.0)}& 48.11 {\scriptsize\color{blue}(+0.0)}& 56.38 {\scriptsize\color{blue}(+0.0)}& 68.09 {\scriptsize\color{blue}(+0.0)}& 53.48 {\scriptsize\color{blue}(+0.0)}& 57.47 {\scriptsize\color{blue}(+0.0)}& 76.85 {\scriptsize\color{blue}(+0.0)}\\
\cmidrule{1-12}\\[-1.56em]
\cmidrule{1-12}
\multirow{24}{*}{\rotatebox{90}{Open-Sourced}} & \multirow{3}{*}{\textbf{LLaVA-1.5-13B}}
& LEAST & 59.74 {\scriptsize\color{red}(-1.86)} & 55.28 {\scriptsize\color{red}(-3.01)} & 31.7 {\scriptsize\color{blue}(+1.51)} & 46.54 {\scriptsize\color{red}(-2.78)} & 47.72 {\scriptsize\color{red}(-5.73)} & 25.94 {\scriptsize\color{red}(-0.14)} & 47.62 {\scriptsize\color{red}(-5.25)} & 48.26 {\scriptsize\color{red}(-5.49)} & 28.69 {\scriptsize\color{red}(-1.16)}\\
& &MORE & 63.64 {\scriptsize\color{blue}(+3.6)} & 53.52 {\scriptsize\color{red}(-4.72)} & 31.99 {\scriptsize\color{blue}(+1.25)} & 46.32 {\scriptsize\color{blue}(+3.53)} & 43.09 {\scriptsize\color{red}(-6.8)} & 25.33 {\scriptsize\color{red}(-1.38)} & 51.6 {\scriptsize\color{blue}(+5.52)} & 44.52 {\scriptsize\color{red}(-6.44)} & 27.96 {\scriptsize\color{red}(-3.41)}\\
& &MOST & 63.16 {\scriptsize\color{red}(-2.21)} & 55.43 {\scriptsize\color{red}(-7.01)} & 64.99 {\scriptsize\color{red}(-11.21)} & 48.05 {\scriptsize\color{red}(-0.06)} & 47.99 {\scriptsize\color{red}(-8.39)} & 62.9 {\scriptsize\color{red}(-5.19)} & 52.64 {\scriptsize\color{red}(-0.84)} & 47.11 {\scriptsize\color{red}(-10.36)} & 72.14 {\scriptsize\color{red}(-4.71)}\\
\cmidrule{2-12}
& \multirow{3}{*}{+C10}
& LEAST & 65.65 {\scriptsize\color{blue}(+4.05)} & 70.43 {\scriptsize\color{blue}(+12.14)} & 64.14 {\scriptsize\color{blue}(+33.95)} & 52.99 {\scriptsize\color{blue}(+3.67)} & 63.24 {\scriptsize\color{blue}(+9.79)} & 50.24 {\scriptsize\color{blue}(+24.16)} & 50.56 {\scriptsize\color{red}(-2.31)} & 59.57 {\scriptsize\color{blue}(+5.82)} & 55.3 {\scriptsize\color{blue}(+25.45)}\\
& &MORE & 68.34 {\scriptsize\color{blue}(+8.3)} & 69.38 {\scriptsize\color{blue}(+11.14)} & 80.14 {\scriptsize\color{blue}(+49.4)} & 55.2 {\scriptsize\color{blue}(+12.41)} & 62.11 {\scriptsize\color{blue}(+12.22)} & 54.48 {\scriptsize\color{blue}(+27.77)} & 54.89 {\scriptsize\color{blue}(+8.81)} & 58.74 {\scriptsize\color{blue}(+7.78)} & 58.86 {\scriptsize\color{blue}(+27.49)}\\
& &MOST & 69.67 {\scriptsize\color{blue}(+4.3)} & 69.69 {\scriptsize\color{blue}(+7.25)} & 81.0 {\scriptsize\color{blue}(+4.8)} & 53.32 {\scriptsize\color{blue}(+5.21)} & 61.59 {\scriptsize\color{blue}(+5.21)} & 72.55 {\scriptsize\color{blue}(+4.46)} & 53.28 {\scriptsize\color{red}(-0.2)} & 58.62 {\scriptsize\color{blue}(+1.15)} & 79.8 {\scriptsize\color{blue}(+2.95)}\\
\cmidrule{2-12}
& \multirow{3}{*}{+TI}
& LEAST & 66.01 {\scriptsize\color{blue}(+4.41)} & 62.02 {\scriptsize\color{blue}(+3.73)} & 68.59 {\scriptsize\color{blue}(+38.4)} & 65.11 {\scriptsize\color{blue}(+15.79)} & 61.71 {\scriptsize\color{blue}(+8.26)} & 71.89 {\scriptsize\color{blue}(+45.81)} & 62.79 {\scriptsize\color{blue}(+9.92)} & 59.41 {\scriptsize\color{blue}(+5.66)} & 74.43 {\scriptsize\color{blue}(+44.58)}\\
& &MORE & 66.41 {\scriptsize\color{blue}(+6.37)} & 64.09 {\scriptsize\color{blue}(+5.85)} & 75.32 {\scriptsize\color{blue}(+44.58)} & 65.31 {\scriptsize\color{blue}(+22.52)} & 61.82 {\scriptsize\color{blue}(+11.93)} & 74.08 {\scriptsize\color{blue}(+47.37)} & 63.43 {\scriptsize\color{blue}(+17.35)} & 59.87 {\scriptsize\color{blue}(+8.91)} & 79.57 {\scriptsize\color{blue}(+48.2)}\\
& &MOST & 66.0 {\scriptsize\color{blue}(+0.63)} & 62.52 {\scriptsize\color{blue}(+0.08)} & 81.19 {\scriptsize\color{blue}(+4.99)} & 64.41 {\scriptsize\color{blue}(+16.3)} & 59.18 {\scriptsize\color{blue}(+2.8)} & 76.17 {\scriptsize\color{blue}(+8.08)} & 62.07 {\scriptsize\color{blue}(+8.59)} & 58.21 {\scriptsize\color{blue}(+0.74)} & 82.04 {\scriptsize\color{blue}(+5.19)}\\
\cmidrule{2-12}
& \multirow{3}{*}{+C10\&TI}
& LEAST & 70.53 {\scriptsize\color{blue}(+8.93)} & 69.1 {\scriptsize\color{blue}(+10.81)} & 57.02 {\scriptsize\color{blue}(+26.83)} & 66.15 {\scriptsize\color{blue}(+16.83)} & 62.47 {\scriptsize\color{blue}(+9.02)} & 62.41 {\scriptsize\color{blue}(+36.33)} & 62.96 {\scriptsize\color{blue}(+10.09)} & 60.86 {\scriptsize\color{blue}(+7.11)} & 64.39 {\scriptsize\color{blue}(+34.54)}\\
& &MORE & 70.05 {\scriptsize\color{blue}(+10.01)} & 68.25 {\scriptsize\color{blue}(+10.01)} & 63.87 {\scriptsize\color{blue}(+33.13)} & 65.99 {\scriptsize\color{blue}(+23.2)} & 63.94 {\scriptsize\color{blue}(+14.05)} & 68.11 {\scriptsize\color{blue}(+41.4)} & 63.58 {\scriptsize\color{blue}(+17.5)} & 62.06 {\scriptsize\color{blue}(+11.1)} & 71.7 {\scriptsize\color{blue}(+40.33)}\\
& &MOST & 69.82 {\scriptsize\color{blue}(+4.45)} & 68.05 {\scriptsize\color{blue}(+5.61)} & 81.46 {\scriptsize\color{blue}(+5.26)} & 64.74 {\scriptsize\color{blue}(+16.63)} & 62.88 {\scriptsize\color{blue}(+6.5)} & 74.73 {\scriptsize\color{blue}(+6.64)} & 62.23 {\scriptsize\color{blue}(+8.75)} & 61.69 {\scriptsize\color{blue}(+4.22)} & 81.3 {\scriptsize\color{blue}(+4.45)}\\
\cmidrule{2-12}\\[-1.56em]
\cmidrule{2-12}
 & \multirow{3}{*}{\textbf{Bunny-1.1-8B}}
& LEAST & 64.37 {\scriptsize\color{blue}(+2.77)} & 58.82 {\scriptsize\color{blue}(+0.53)} & 27.36 {\scriptsize\color{red}(-2.83)} & 51.98 {\scriptsize\color{blue}(+2.66)} & 52.3 {\scriptsize\color{red}(-1.15)} & 27.43 {\scriptsize\color{blue}(+1.35)} & 53.64 {\scriptsize\color{blue}(+0.77)} & 52.87 {\scriptsize\color{red}(-0.88)} & 30.39 {\scriptsize\color{blue}(+0.54)}\\
& &MORE & 65.27 {\scriptsize\color{blue}(+5.23)} & 53.73 {\scriptsize\color{red}(-4.51)} & 26.79 {\scriptsize\color{red}(-3.95)} & 49.46 {\scriptsize\color{blue}(+6.67)} & 45.24 {\scriptsize\color{red}(-4.65)} & 25.85 {\scriptsize\color{red}(-0.86)} & 53.32 {\scriptsize\color{blue}(+7.24)} & 45.21 {\scriptsize\color{red}(-5.75)} & 28.79 {\scriptsize\color{red}(-2.58)}\\
& &MOST & 68.89 {\scriptsize\color{blue}(+3.52)} & 60.38 {\scriptsize\color{red}(-2.06)} & 62.69 {\scriptsize\color{red}(-13.51)} & 52.94 {\scriptsize\color{blue}(+4.83)} & 54.12 {\scriptsize\color{red}(-2.26)} & 58.15 {\scriptsize\color{red}(-9.94)} & 57.35 {\scriptsize\color{blue}(+3.87)} & 51.75 {\scriptsize\color{red}(-5.72)} & 74.77 {\scriptsize\color{red}(-2.08)}\\
\cmidrule{2-12}
& \multirow{3}{*}{+C10}
& LEAST & 68.86 {\scriptsize\color{blue}(+7.26)} & 66.73 {\scriptsize\color{blue}(+8.44)} & 25.98 {\scriptsize\color{red}(-4.21)} & 56.01 {\scriptsize\color{blue}(+6.69)} & 59.47 {\scriptsize\color{blue}(+6.02)} & 26.78 {\scriptsize\color{blue}(+0.7)} & 55.09 {\scriptsize\color{blue}(+2.22)} & 55.15 {\scriptsize\color{blue}(+1.4)} & 29.33 {\scriptsize\color{red}(-0.52)}\\
& &MORE & 69.86 {\scriptsize\color{blue}(+9.82)} & 64.54 {\scriptsize\color{blue}(+6.3)} & 25.85 {\scriptsize\color{red}(-4.89)} & 53.55 {\scriptsize\color{blue}(+10.76)} & 52.38 {\scriptsize\color{blue}(+2.49)} & 25.96 {\scriptsize\color{red}(-0.75)} & 54.44 {\scriptsize\color{blue}(+8.36)} & 48.2 {\scriptsize\color{red}(-2.76)} & 27.59 {\scriptsize\color{red}(-3.78)}\\
& &MOST & 69.98 {\scriptsize\color{blue}(+4.61)} & 67.55 {\scriptsize\color{blue}(+5.11)} & 66.14 {\scriptsize\color{red}(-10.06)} & 55.31 {\scriptsize\color{blue}(+7.2)} & 58.81 {\scriptsize\color{blue}(+2.43)} & 59.33 {\scriptsize\color{red}(-8.76)} & 58.04 {\scriptsize\color{blue}(+4.56)} & 54.18 {\scriptsize\color{red}(-3.29)} & 73.71 {\scriptsize\color{red}(-3.14)}\\
\cmidrule{2-12}
& \multirow{3}{*}{+TI}
& LEAST & 69.3 {\scriptsize\color{blue}(+7.7)} & 65.75 {\scriptsize\color{blue}(+7.46)} & 26.37 {\scriptsize\color{red}(-3.82)} & 59.83 {\scriptsize\color{blue}(+10.51)} & 60.42 {\scriptsize\color{blue}(+6.97)} & 29.26 {\scriptsize\color{blue}(+3.18)} & 56.67 {\scriptsize\color{blue}(+3.8)} & 55.76 {\scriptsize\color{blue}(+2.01)} & 31.58 {\scriptsize\color{blue}(+1.73)}\\
& &MORE & 69.65 {\scriptsize\color{blue}(+9.61)} & 62.65 {\scriptsize\color{blue}(+4.41)} & 26.39 {\scriptsize\color{red}(-4.35)} & 58.83 {\scriptsize\color{blue}(+16.04)} & 56.51 {\scriptsize\color{blue}(+6.62)} & 27.83 {\scriptsize\color{blue}(+1.12)} & 54.98 {\scriptsize\color{blue}(+8.9)} & 48.37 {\scriptsize\color{red}(-2.59)} & 30.62 {\scriptsize\color{red}(-0.75)}\\
& &MOST & 69.51 {\scriptsize\color{blue}(+4.14)} & 66.63 {\scriptsize\color{blue}(+4.19)} & 62.23 {\scriptsize\color{red}(-13.97)} & 57.69 {\scriptsize\color{blue}(+9.58)} & 62.4 {\scriptsize\color{blue}(+6.02)} & 57.56 {\scriptsize\color{red}(-10.53)} & 58.09 {\scriptsize\color{blue}(+4.61)} & 54.77 {\scriptsize\color{red}(-2.7)} & 72.67 {\scriptsize\color{red}(-4.18)}\\
\cmidrule{2-12}
& \multirow{3}{*}{+C10\&TI}
& LEAST & 69.81 {\scriptsize\color{blue}(+8.21)} & 67.14 {\scriptsize\color{blue}(+8.85)} & 27.52 {\scriptsize\color{red}(-2.67)} & 60.13 {\scriptsize\color{blue}(+10.81)} & 60.43 {\scriptsize\color{blue}(+6.98)} & 30.02 {\scriptsize\color{blue}(+3.94)} & 56.94 {\scriptsize\color{blue}(+4.07)} & 55.4 {\scriptsize\color{blue}(+1.65)} & 33.06 {\scriptsize\color{blue}(+3.21)}\\
& &MORE & 70.2 {\scriptsize\color{blue}(+10.16)} & 63.0 {\scriptsize\color{blue}(+4.76)} & 28.05 {\scriptsize\color{red}(-2.69)} & 58.7 {\scriptsize\color{blue}(+15.91)} & 54.76 {\scriptsize\color{blue}(+4.87)} & 27.99 {\scriptsize\color{blue}(+1.28)} & 55.14 {\scriptsize\color{blue}(+9.06)} & 48.56 {\scriptsize\color{red}(-2.4)} & 31.48 {\scriptsize\color{blue}(+0.11)}\\
& &MOST & 70.13 {\scriptsize\color{blue}(+4.76)} & 69.46 {\scriptsize\color{blue}(+7.02)} & 68.68 {\scriptsize\color{red}(-7.52)} & 58.27 {\scriptsize\color{blue}(+10.16)} & 63.12 {\scriptsize\color{blue}(+6.74)} & 61.51 {\scriptsize\color{red}(-6.58)} & 58.42 {\scriptsize\color{blue}(+4.94)} & 55.9 {\scriptsize\color{red}(-1.57)} & 74.92 {\scriptsize\color{red}(-1.93)}\\
\bottomrule
\end{tabular}
\end{adjustbox}\vspace{-2mm}
\caption{\coin-Bench Results (\%). Small-scale \coin-Instruct-finetuned opensource VLMs can outperform GPT-4o. (.) shows the accuracy \textcolor{blue}{gain}/\textcolor{red}{gap} to GPT-4o.}\vspace{-2mm}
\label{tab:coin_bench}
\end{table*}

\subsection{Results}
\subsubsection{\coin-Bench Results}

\paragraph{Comparison with GPT4o} 
GPT-4o remains one of the best closed-source model in many aspects for its massive and surperior training data and method. However, we find that fine-tuning on \coin-Instruct significantly improved the accuracy of these models across nearly all tasks on \coin-Bench, \textit{some even surpass GPT-4o}. LLaVA-1.5 finetuned with \coin-Instruct, in particular, consistently outperforms GPT-4o, even in the more challenging splits like ImageNet, showcasing the effectiveness of \coin-Instruct in enhancing model performance. The fine-tuned Bunny-1.1 also outperformed GPT-4o in most tasks, although we observed a notable exception with the C10ENV task, where Bunny-1.1's accuracy was 2.67\% to 10.06\% lower than GPT-4o. Bunny-1.1 also exhibited slightly lower performance on the ImageNet dataset, which was not included in its training set, suggesting potential limitations in generalization to new domains. We want to reiterate that LLaVA-1.5 we use is a 13B model, and Bunny-1.1 being a 8B model, which is massively smaller than closed-source models and without full-finetuning.

The results show that our SFT dataset and benchmark play a vital role in enhancing model performance in the visual perception tasks. Our structured prompts and varied domains in our benchmark allowed LLaVA-1.5 to consistently outperform GPT-4o, showcasing the effectiveness of targeted fine-tuning. The occasional performance drop for C10ENV and Bunny-1.1 on unseen tasks in ImageNet highlights areas where broader generalization may be limited. Our dataset and benchmark effectively push models beyond the capabilities seen in zero-shot settings, providing a meaningful boost in domain-specific performance that zero-shot models like GPT-4o could not achieve.

\vspace{-2mm}
\paragraph{Open-source VLMs}
On LLaVA-1.5 and Bunny-1.1, we observed performance improvement when models were fine-tuned and tested within the same domain (Tab.~\ref{tab:coin_bench}). For instance, LLaVA-1.5 demonstrates notable gains on the LEAST instruction within TIFGD, with accuracy rising from 46.54\% to 65.11\% after fine-tuning with TI. This improvement illustrates that our dataset plays a positive role in data augmentation on the color and environment recognition tasks.

Additionally, we notice out-of-distribution domain adaptation capabilities using \coin-Instruct, with performance gains even on previously unseen data. LLaVA-1.5, for instance, shows an increase on TIENV under the MOST prompt from 62.9\% to 72.55\% after fine-tuning on the C10 split of \coin-Instruct. Similarly, LLaVA-1.5's accuracy on INetBGD with MORE prompts increased from 44.52\% to 62.06\% when augmented with C10+TI, despite the INet dataset being both unseen and of higher resolution than C10+TI.

Another interesting observation is that while a notable performance improvement is evident from the LEAST to the MOST prompt especially on ENV, the MORE prompt does not consistently benefit performance. In some cases, it even leads to performance drops, even on GPT-4o, indicating that the model may struggle to identify the target object in the image. This reveals the design of our benchmark introduces specific challenges to the models.

These findings highlight how \coin-Bench's Tiered-MQA design enables uncovering models' ability to improve specific tasks and \coin-Instruct can benefit domain adaptation.

\vspace{-2mm}
\paragraph{Qualitative Results}
\begin{figure*}
    \centering
    \includegraphics[width=\textwidth]{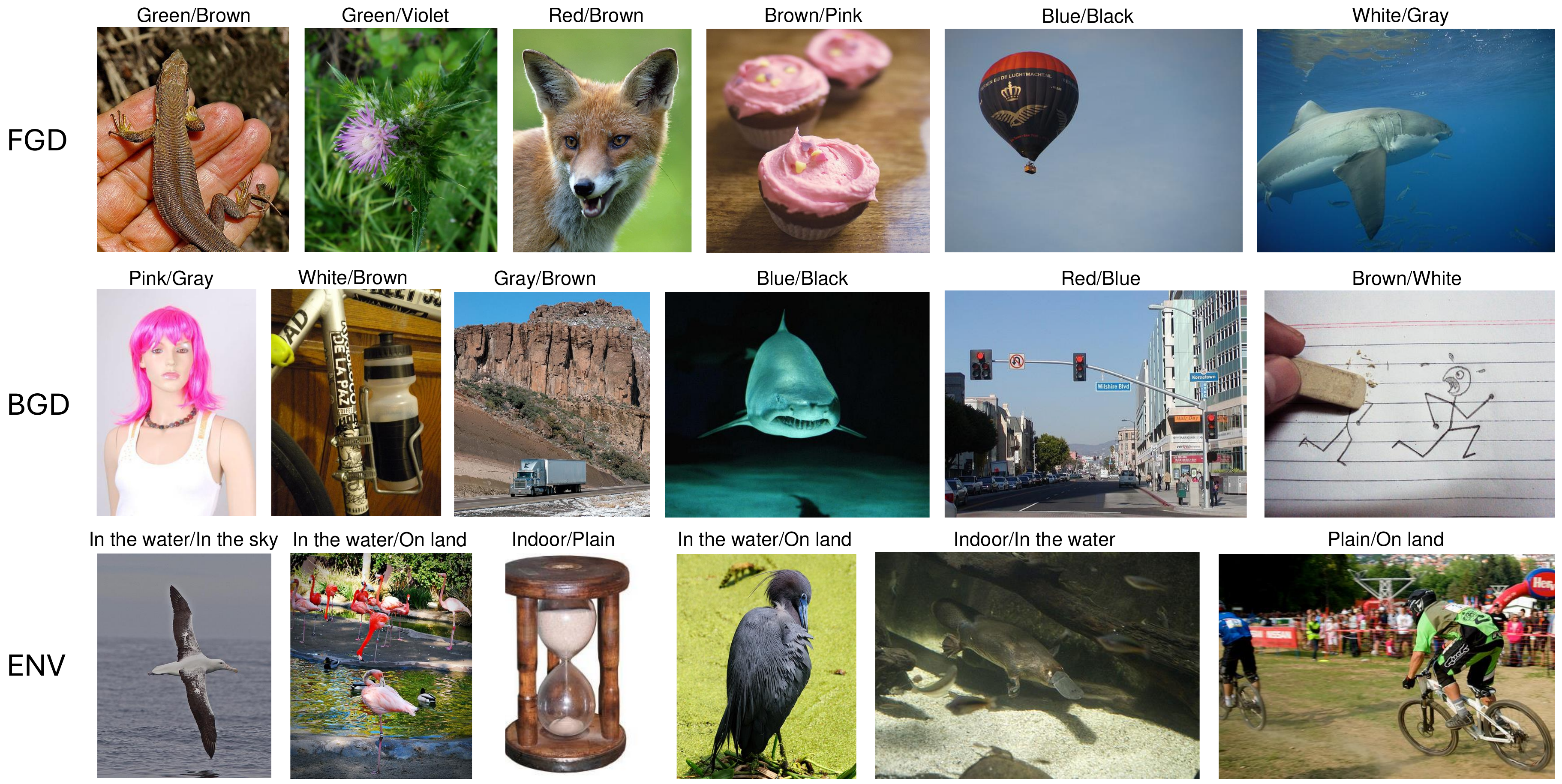}\vspace{-2mm}
    \caption{Failure cases on \coin-Bench before/after training with \coin-Instruct. After fine-tuning with \coin-Instruct, we are able to have the VLMs recognize the correct colors and environments.}
    \label{fig:failure}\vspace{-5mm}
\end{figure*}
In Fig.~\ref{fig:failure}, we show qualitative examples, where using \coin-Instruct helps correct wrong perception, to showcase how our \coin-Instruct improves VLM performance on \coin-Bench. (1) \textit{FGD misperception}: in an image of a ``Great White Shark" (row 1, 6th image), where the VLM may initially answer ``white" based on the species name, which is a contextual bias. After \coin-Instruct, it correctly answers ``gray," reflecting the actual color. Similar biases occur with images of a “Green Lizard” and ``Red Fox," where the model may rely on species names instead of their true colors. (2) \textit{BGD misinterpretation:} in a traffic light image (row 2, 5th image), the VLM misinterpreted the background as ``red," which may be due to it focusing on the light’s color instead of the surroundings. After training with \coin-instruct, the VLM accurately identifies the background as ``blue," resolving object-background distinctions. (3) \textit{ENV mis-judgement:} an image of a seagull flying above the sea (row 3, 1st image) have the VLM answer ``in the water." We suspect that it is because the background confuses the model, but after training with \coin-Instruct, it correctly identifies the environment as ``in the sky." 

These qualitative examples reveal the usefulness of \coin-Instruct in enhancing the VLM ability in basic vision tasks and how \coin-Bench can facilitate such analyses of VLMs. 

\vspace{-2mm}
\subsubsection{Common Benchmark Results}
\vspace{-2mm}
\begin{table}[t]
    \centering
    \begin{adjustbox}{width=\linewidth}

    \begin{tabular}{lcccccc}
    \toprule
         & MMBench &  MME-P & MME-C & VQAv2 & GQA \\ \hline
         LLaVA-1.5-13B  & 67.70 &  1530.43 & 298.57& 80 & 63.3\\
      +C10 & \textbf{68.21}  &\textbf{1530.91} & 276.78& \textbf{80.07} & \textbf{63.4}\\
      +TI &67.35 & \textbf{1561.84} & 294.28 & \textbf{80.18}& \textbf{63.4} \\
      +C10\&TI & 66.41 & \textbf{1538.36} & 276.79 & \textbf{80.22} & \textbf{63.49}\\ 
      +220k & 66.41 & \textbf{1574.82} & 279.64 & \textbf{80.16} & \textbf{63.49}\\
      \midrule
      Bunny-1.1-8B  & 76.63 &  1647.81 & 305.36 & 82.51 & 64.25\\
    +C10 & \textbf{77.06} &  1639.08 & \textbf{344.29} & 82.35 & \textbf{64.46} \\
      +TI & \textbf{76.63} & \textbf{1651.93} & \textbf{332.14} & \textbf{82.52} & 64.18 \\
      +C10\&TI & 76.03  & 1646.26 & \textbf{337.86} & \textbf{82.65} & 64.01 \\
      +220k & \textbf{78.44}  & 1630.19 & \textbf{352.14} & \textbf{82.65} & \textbf{64.47} \\
      \bottomrule
    \end{tabular}
    \end{adjustbox}\vspace{-2mm}
    \caption{Performance of VLMs on common benchmarks.}\vspace{-2mm}
    \label{tab:vlm_bench}
\end{table}


         
Tab.~\ref{tab:vlm_bench} shows the SFT results of our fine-tuned models on common benchmarks such as MMBench, MME, VQAv2, and GQA. In addition to the four variants mentioned in Sec.~\ref{sec:exp_setup}, we also explore fine-tuning the VLMs with combined original SFT data and the entire \coin~(dubbed 220k) because they are not evaluated on ImageNet. We suspected fine-tuning LLaVA-1.5 and Bunny-1.1 VLMs with \coin~would lead to an overall performance drop as \coin~focuses on more basic visual capabilities. Surprisingly, we observe Bunny and LLaVA show improvements on more than half of the cases; and on VQAv2 and GQA, fine-tuned models is either on par with or slightly better than the baselines. These results suggest that fine-tuning with our dataset would lead to performance gains in the majority cases, despite a few moderate performance drops on specific model-data combinations. It further corroborates the value of \coin.

\begin{table*}
\centering
\begin{adjustbox}{width=\textwidth}
\begin{tabular}{lrrrrrrrccc}
\toprule
\textbf{Dataset} &  \textbf{\# Attr.} & \textbf{\# Classes} & \textbf{\# Train set} & \textbf{\# Val. set} & \textbf{\# Test set} & \textbf{Max group} & \textbf{Min group} & \textbf{SC} & \textbf{AI} & \textbf{AG} \\ \midrule
\textbf{Waterbirds}   & 2  & 2  & 4795  & 1199  & 5794  & 3498 (73.0\%)  & 56 (1.2\%)  & x & x &  \\ 
\textbf{CelebA}   & 2  & 2  & 162770 & 19867 & 19962 & 71629 (44.0\%)  & 1387 (0.9\%)  & x &  & \\ 
\textbf{MetaShift}  & 2  & 2  & 2276  & 349   & 874   & 789 (34.7\%)  & 196 (8.6\%)   & x & & \\ 
\textbf{Spawrious (O2O-Easy setup)}    & 6 & 4 & 12672 & n/a & 12672 &  3073 (24.25\%)& 3802 (3\%) & x & x &  \\ 

\textbf{\coin-C10-FGD}& 12 & 10 & 50000 & 10000 & n/a & 15734 (26.2\%) & 232 (0.4\%)  & x & x & x \\
\textbf{\coin-C10-FGD-\textit{merge}}& 12 & 10 & 50000 & 10000 & n/a & 15734 (26.2\%) & 232 (0.4\%)  & x & x & x \\ 
\textbf{\coin-TI-FGD} & 12 & 200 & 100000 & 10000 & n/a & 25113 (22.8\%) & 1036 (0.9\%) & x & x & x \\ 
\textbf{\coin-TI-FGD-\textit{merge}} & 12 & 200 & 100000 & 10000 & n/a & 25113 (22.8\%) & 5383 (4.9\%) & x & x & x \\
\textbf{\coin-C10-BGD} & 12 & 10 & 50000 & 10000 & n/a & 15599 (26.0\%) & 329 (0.5\%) & x & x & x \\ 
\textbf{\coin-C10-BGD-\textit{merge}} & 12 & 10 & 50000 & 10000 & n/a & 15599 (26.0\%) & 759 (1.3\%) & x & x & x \\
\textbf{\coin-TI-BGD} & 12 & 200 & 100000 & 10000 & n/a & 33923 (30.8\%) & 638 (0.6\%) & x & x & x \\ 
\textbf{\coin-TI-BGD-\textit{merge}} & 12 & 200 & 100000 & 10000 & n/a & 33923 (30.8\%) & 2552 (2.3\%) & x & x & x \\ 
\textbf{\coin-C10-ENV} & 6 & 10 & 50000 & 10000 & n/a & 43528 (72.5\%) & 1373 (2.3\%) & x & x & x \\ 
\textbf{\coin-TI-ENV} & 6 & 200 & 100000 & 10000 & n/a & 62311 (56.6\%) & 681 (0.6\%) & x & x & x \\ 
\bottomrule
\end{tabular}
\end{adjustbox}\vspace{-2mm}
\caption{Common image domain generalization datasets. \coin~imposes more challenge in terms of more attributes and more classes. Also, \coin~supports all SC, AI, AG (see Sec.~\ref{sec:relatedworks}) evaluation scenarios.}\vspace{-2mm}
\label{table:DG_datasets_comparison}
\end{table*}

\section{Domain Generalization}\label{sec:DG}
Our medium-grained contextual labels in \coin~can serve as \textit{spurious} attributes, suitable for benchmarking DG algorithms along with a few other desiderata mentioned in Sec.~\ref{sec:relatedworks}. We evaluate the domain generalization algorithms in the linear probing setting of VLM in this section. We also provide the training from scratch Resnet18 results in the Appendix for completeness. 
\vspace{-4mm}
\paragraph{Setup} We adopt the implementation of DomainBed \cite{gulrajani2020searchlostdomaingeneralization} using the CLIP representation \cite{radford2021learningtransferablevisualmodels} and do 10 runs each with different random seeds of seven different DG algorithms, which are ERM, GroupDRO \cite{sagawa2019distributionally}, CORAL\cite{sun2016deep}, MMD\cite{li2018domain}, EQRM\cite{eastwood2022probable}, SelfReg\cite{kim2021selfreg} and VREx\cite{krueger2021out} based upon their performances in previous studies \cite{lynch2023spawrious, gulrajani2020searchlostdomaingeneralization} and recency. 
\vspace{-4mm}
\paragraph{Merging similar colors} Some subgroups have way fewer instances while being similar in hue, such as violet and blue, so we also explore a \textit{merge} version of \coin by merging pink to red, orange to yellow, and violet to blue to reduce the complexities of our dataset. 

\begin{table*}[t]
\centering
\begin{adjustbox}{width=\textwidth}
\begin{tabular}{lcccccccccc|c}
\toprule
Method  & C10FGD 
&C10FGDmerge 
& C10BGD & C10BGDmerge & C10ENV  & TIFGD 
&TIFGDmerge 
& TIBGD & TIBGDmerge & TIENV & Average \\ \midrule
ERM  & 97.68 (0.08) 
&97.62 (0.13) 
& 97.68 (0.15) & 97.72 (0.17) & 97.58 (0.12)  & 80.18 (0.19) 
&80.63 (0.39) 
& 80.73 (0.34) & 80.73 (0.29) & 79.89 (0.25) & \underline{89.04} (0.21) \\
GroupDRO  & 97.61 (0.14) 
&97.60 (0.08) 
& 97.69 (0.16) & 97.75 (0.09) & 97.51 (0.12)  & 79.42 (0.45) 
&79.92 (0.31) 
& 79.53 (0.41) & 80.03 (0.60) & 78.54 (0.83) & 88.56 (0.32) \\
CORAL  & 97.63 (0.10) 
&97.64 (0.10) 
& 97.74 (0.10) & 97.69 (0.11) & 97.62 (0.07)  & 80.37 (0.24) 
&80.51 (0.32) 
& 80.59 (0.38) & 80.78 (0.29) & 79.79 (0.37) & \underline{89.04} (0.21) \\
EQRM  & 97.62 (0.10) 
&97.69 (0.10) 
& 97.66 (0.16) & 97.70 (0.11) & 97.54 (0.13)  & 80.11 (0.31) 
&80.35 (0.32) 
& 80.31 (0.32) & 80.56 (0.23) & 79.59 (0.30) & 88.91 (0.21) \\
MMD  & 97.60 (0.07) 
&97.64 (0.12) 
& 97.83 (0.07) & 97.78 (0.10) & 97.57 (0.12)  & 80.35 (0.21) 
&80.52 (0.42) 
& 80.54 (0.19) & 80.90 (0.31) & 79.82 (0.33) & \textbf{89.06} (0.19) \\
SelfReg  & 97.50 (0.09) 
&97.51 (0.13) 
& 97.60 (0.08) & 97.58 (0.13) & 97.51 (0.09)  & 79.70 (0.29) 
&79.60 (0.25) 
& 79.78 (0.25) & 80.03 (0.32) & 79.03 (0.38) & 88.58 (0.20) \\
VREx  & 97.44 (0.29) 
&97.52 (0.20) 
& 97.65 (0.15) & 97.58 (0.16) & 97.50 (0.14)  & 77.60 (2.29) 
&77.85 (2.66) 
& 77.82 (2.00) & 77.34 (3.10) & 78.05 (1.62) & 87.64 (1.26) \\
\midrule
Average  & 97.58 (0.12) &97.60 (0.12) & 97.69 (0.12) & 97.69 (0.12) & 97.55 (0.11)  & 79.68 (0.57) &79.91 (0.67) & 79.98 (0.56) & 80.05 (0.73) & 79.24 (0.58) & 88.69 (0.37)\\
\bottomrule
\end{tabular}
\end{adjustbox}\vspace{-2mm}
\caption{Domain generalization algorithms results (\%, std) on \coin~in linear probing. We observe that MMD is the overall best, immediately followed by ERM and CORAL.}\vspace{-2mm}
\label{tab:transposed_linear_probing}
\end{table*}

\subsection{Result}

\paragraph{MMD is the best on generalization, followed by CORAL and ERM.} The generalization results across all datasets (Tab.~\ref{tab:transposed_linear_probing}) indicate that MMD achieves the highest average accuracy, at 89.06\% with the lowest standard deviation (0.19), indicating its robust generalization across different dataset domains. ERM and CORAL follow closely with similar average accuracy (both at 89.04\%) but slightly higher variance compared to MMD, suggesting slight sensitivity. GroupDRO and SelfReg maintain competitive performance levels, though consistently lower than MMD and CORAL, while VREx displays both the lowest accuracy (87.64\%) and the highest variability (1.26), indicating less stability in its performance.

\vspace{-2mm}
\paragraph{CORAL is the best with subpopulation shift.}

\begin{table}[h]
\centering
\begin{adjustbox}{width=\linewidth}
\begin{tabular}{cccccccc|c}
\toprule
Trained$\times$Evaluated & ERM &GroupDRO & CORAL & EQRM & MMD & SelfReg & VREx & Average\\ 
\midrule
C10FGD$\times$FGD  & 95.00 &95.24 & 96.19 & 95.00 & 95.00 & 93.57 & 95.00  & 95.00\\
C10BGD$\times$BGD  & 94.68 &93.94 & 94.68 & 94.47 & 94.79 & 94.57 & 93.94  & 94.44\\
C10ENV$\times$ENV  & 96.03 &95.94 & 96.23 & 95.88 & 96.05 & 96.01 & 95.98  & 96.02\\
TIFGD$\times$FGD  & 75.05 &75.62 & 76.48 & 76.38 & 74.95 & 75.81 & 72.29  & 75.23\\
TIBGD$\times$BGD  & 75.40 &74.92 & 75.81 & 75.08 & 75.81 & 75.56 & 71.85  & 74.92\\
TIENV$\times$ENV  & 76.97 &75.11 & 76.65 & 76.73 & 77.53 & 76.16 & 74.99  & 76.31\\
\midrule
Average & 85.21 & 85.13 & \textbf{86.01} & 85.59 & \underline{85.69} & 85.28
 & 84.01& 85.32\\\bottomrule
\end{tabular}
\end{adjustbox}\vspace{-2mm}
\caption{Worst group accuracy (\%) when using the same training and evaluation subgroups.}
\label{tab:worst_group_self}\vspace{-2mm}
\end{table}
\begin{table}[h]
\centering
\begin{adjustbox}{width=\linewidth}
\begin{tabular}{ccccccccc|c}
\toprule
\multicolumn{2}{l}{Trained $\times$ Evaluated} & ERM  &GroupDRO & CORAL & EQRM & MMD & SelfReg & VREx & Average\\ \midrule
\multirow{5}{*}{FGD} & C10FGD$\times$BGD & 93.19  &94.36 & 94.47 & 94.36 & 94.47 & 93.72 & 94.04 & 94.09\\ 
& C10FGD$\times$ENV & 96.27  &95.46 & 95.95 & 95.94 & 95.81 & 96.07 & 95.41 & 95.84\\ 
& TIFGD$\times$BGD & 74.52  &73.31 & 74.19 & 73.63 & 73.79 & 73.79 & 69.76 & 73.28\\ 
& TIFGD$\times$ENV & 77.34  &76.43 & 77.87 & 77.43 & 77.41 & 77.23 & 74.47 & 76.88\\
\cmidrule{2-10}
& Average & \underline{85.33}  &84.89 & \textbf{85.62} & 85.34 & 85.37 & 85.20 & 83.42 & 85.02\\
\midrule
\multirow{5}{*}{BGD} & C10BGD$\times$FGD & 96.37  &96.44 & 95.95 & 95.00 & 96.76 & 93.81 & 96.45 & 95.83\\ 
& C10BGD$\times$ENV & 96.17  &96.28 & 96.37 & 96.04 & 96.42 & 95.98 & 95.88 & 96.16\\ 
& TIBGD$\times$FGD & 76.95  &75.90 & 77.52 & 76.76 & 75.14 & 76.48 & 71.90 & 75.81\\ 
& TIBGD$\times$ENV & 78.02  &76.26 & 77.53 & 77.70 & 77.72 & 77.21 & 75.26 & 77.10\\
\cmidrule{2-10}
&Average & \textbf{86.88}  &86.22 & \underline{86.84} & 86.38 & 86.51 & 85.87 & 84.87& 86.22\\
\midrule
\multirow{5}{*}{ENV}& C10ENV$\times$FGD & 95.95  &96.32 & 95.71 & 95.71 & 96.43 & 94.52 & 94.52 & 95.59\\ 
& C10ENV$\times$BGD & 94.04  &93.62 & 94.26 & 94.04 & 93.40 & 94.63 & 94.36 & 94.05\\ 
& TIENV$\times$FGD & 76.06  &73.24 & 76.12 & 75.91 & 75.97 & 75.56 & 74.87 & 75.39\\ 
& TIENV$\times$BGD & 76.85  &74.03 & 75.56 & 75.89 & 75.56 & 75.00 & 73.79 & 75.24\\
\cmidrule{2-10}
&Average & \textbf{85.73}  &84.30 & \underline{85.41} & 85.39 & 85.34 & 84.93 & 84.39 & 85.07\\
\bottomrule
\end{tabular}
\end{adjustbox}\vspace{-2mm}
\caption{Worst group performance (\%) in attribute generalization (AG) setting. Contrary to the observations from generalization performance, CORAL and ERM excel in the AG setting, but MMD remain competitive.}
\label{tab:worst_group}\vspace{-5mm}
\end{table}

Subpopulation shift is a branch in DG and uses worst-group accuracy as the gold-standard metric. Tab.~\ref{tab:worst_group_self} presents the WGA that are evaluated on the same set of subgroups as training. We surprisingly observe that the \textit{subgroup robust} method like GroupDRO was not among the top performers, meaning there is room for improvement this type of method in the linear probing setup. The CORAL still outperforms the others by at least an average 0.32\%, followed by MMD. And VREx still cannot perform at the same level, trailing the second worst by an average of 1.12\%.
\vspace{-2mm}
\paragraph{ERM excels at attribute generalization.} Tab.~\ref{tab:worst_group} shows the WGA of algorithms trained with a certain set of attribute present, but evaluated using another set of attribute, meaning the subgroups used for evaluation are missing during training. We observe the vanilla ERM is the strongest, followed by CORAL and EQRM, contrary to MMD leading the way in the previous evaluations. 


\section{Conclusion}
In this work, we constructed \coin, a large-scale human-annotated dataset containing 660k meduim-grained annotations that can be purposed into a 450k visual instruction tuning dataset (\coin-Instruct) and a 210k evaluation set (\coin-Bench), covering basic visual perception tasks. With \coin-Bench and our Tiered-MQA design, we revealed current VLMs have room to improve on basic visual understanding, and we investigated \coin-Instruct's effectiveness in improving VLM perception capabilities. In addition, due to the medium-grained labels allowing for subgrouping, we investigated \coin's other usage in terms of benchmarking domain generalization methods and found CORAL, ERM, and MMD excel in different evaluation metrics. We hope that our work can help to enhance the alignment from the upstream basic perception task and serve as a benchmark for domain generalization algorithms. Future work on VLM alignment can explore replacing the template instruction pairs with more powerful VLMs to generate guided instruction pairs to enhance the quality of \coin. Another promising line of work can integrate \coin~with noisy data \cite{hendrycks2019robustness} for both VLM alignment and benchmarking algorithms focusing on out-of-distribution generalization. 


{
    \small
    \bibliographystyle{ieeenat_fullname}
    \bibliography{main}

\begin{thebibliography}{41}
\providecommand{\natexlab}[1]{#1}
\providecommand{\url}[1]{\texttt{#1}}
\expandafter\ifx\csname urlstyle\endcsname\relax
  \providecommand{\doi}[1]{doi: #1}\else
  \providecommand{\doi}{doi: \begingroup \urlstyle{rm}\Url}\fi

\bibitem[Achiam et~al.(2023)Achiam, Adler, Agarwal, Ahmad, Akkaya, Aleman, Almeida, Altenschmidt, Altman, Anadkat, et~al.]{achiam2023gpt}
Josh Achiam, Steven Adler, Sandhini Agarwal, Lama Ahmad, Ilge Akkaya, Florencia~Leoni Aleman, Diogo Almeida, Janko Altenschmidt, Sam Altman, Shyamal Anadkat, et~al.
\newblock Gpt-4 technical report.
\newblock \emph{arXiv preprint arXiv:2303.08774}, 2023.

\bibitem[Antol et~al.(2015)Antol, Agrawal, Lu, Mitchell, Batra, Zitnick, and Parikh]{antol2015vqa}
Stanislaw Antol, Aishwarya Agrawal, Jiasen Lu, Margaret Mitchell, Dhruv Batra, C~Lawrence Zitnick, and Devi Parikh.
\newblock Vqa: Visual question answering.
\newblock In \emph{Proceedings of the IEEE international conference on computer vision}, pages 2425--2433, 2015.

\bibitem[Arjovsky et~al.(2019)Arjovsky, Bottou, Gulrajani, and Lopez-Paz]{arjovsky2019invariant}
Martin Arjovsky, L{\'e}on Bottou, Ishaan Gulrajani, and David Lopez-Paz.
\newblock Invariant risk minimization.
\newblock \emph{arXiv preprint arXiv:1907.02893}, 2019.

\bibitem[Beery et~al.(2020)Beery, Cole, and Gjoka]{beery2020iwildcam}
Sara Beery, Elijah Cole, and Arvi Gjoka.
\newblock The iwildcam 2020 competition dataset.
\newblock \emph{arXiv preprint arXiv:2004.10340}, 2020.

\bibitem[Chiu et~al.(2022)Chiu, Wang, Kim, Chen, and Ma]{chiu2022human}
Ming-Chang Chiu, Yingfei Wang, Derrick Eui~Gyu Kim, Pin-Yu Chen, and Xuezhe Ma.
\newblock On human visual contrast sensitivity and machine vision robustness: A comparative study.
\newblock \emph{arXiv preprint arXiv:2212.08650}, 2022.

\bibitem[Chiu et~al.(2023)Chiu, Chen, and Ma]{chiu2023better}
Ming-Chang Chiu, Pin-Yu Chen, and Xuezhe Ma.
\newblock Better may not be fairer: A study on subgroup discrepancy in image classification.
\newblock In \emph{Proceedings of the IEEE/CVF International Conference on Computer Vision}, pages 4956--4966, 2023.

\bibitem[Eastwood et~al.(2022)Eastwood, Robey, Singh, Von~K{\"u}gelgen, Hassani, Pappas, and Sch{\"o}lkopf]{eastwood2022probable}
Cian Eastwood, Alexander Robey, Shashank Singh, Julius Von~K{\"u}gelgen, Hamed Hassani, George~J Pappas, and Bernhard Sch{\"o}lkopf.
\newblock Probable domain generalization via quantile risk minimization.
\newblock \emph{Advances in Neural Information Processing Systems}, 35:\penalty0 17340--17358, 2022.

\bibitem[Fu et~al.(2024)Fu, Chen, Shen, Qin, Zhang, Lin, Yang, Zheng, Li, Sun, Wu, and Ji]{fu2024mmecomprehensiveevaluationbenchmark}
Chaoyou Fu, Peixian Chen, Yunhang Shen, Yulei Qin, Mengdan Zhang, Xu Lin, Jinrui Yang, Xiawu Zheng, Ke Li, Xing Sun, Yunsheng Wu, and Rongrong Ji.
\newblock Mme: A comprehensive evaluation benchmark for multimodal large language models, 2024.

\bibitem[Furness et~al.(2003)Furness, Connor, Robinson, Norton, Ameratunga, and Jackson]{furness2003car}
S Furness, J Connor, Elizabeth Robinson, R Norton, S Ameratunga, and Rodger Jackson.
\newblock Car colour and risk of car crash injury: population based case control study.
\newblock \emph{Bmj}, 327\penalty0 (7429):\penalty0 1455--1456, 2003.

\bibitem[Gulrajani and Lopez-Paz(2020)]{gulrajani2020searchlostdomaingeneralization}
Ishaan Gulrajani and David Lopez-Paz.
\newblock In search of lost domain generalization, 2020.

\bibitem[He et~al.(2024)He, Liu, Wu, Yuan, Wang, Huang, and Zhao]{he2024efficient}
Muyang He, Yexin Liu, Boya Wu, Jianhao Yuan, Yueze Wang, Tiejun Huang, and Bo Zhao.
\newblock Efficient multimodal learning from data-centric perspective.
\newblock \emph{arXiv preprint arXiv:2402.11530}, 2024.

\bibitem[Hendrycks and Dietterich(2019)]{hendrycks2019robustness}
Dan Hendrycks and Thomas Dietterich.
\newblock Benchmarking neural network robustness to common corruptions and perturbations.
\newblock \emph{Proceedings of the International Conference on Learning Representations}, 2019.

\bibitem[Hu et~al.(2021)Hu, Shen, Wallis, Allen-Zhu, Li, Wang, Wang, and Chen]{hu2021lora}
Edward~J Hu, Yelong Shen, Phillip Wallis, Zeyuan Allen-Zhu, Yuanzhi Li, Shean Wang, Lu Wang, and Weizhu Chen.
\newblock Lora: Low-rank adaptation of large language models.
\newblock \emph{arXiv preprint arXiv:2106.09685}, 2021.

\bibitem[Hudson and Manning(2019)]{hudson2019gqa}
Drew~A Hudson and Christopher~D Manning.
\newblock Gqa: A new dataset for real-world visual reasoning and compositional question answering.
\newblock In \emph{Proceedings of the IEEE/CVF conference on computer vision and pattern recognition}, pages 6700--6709, 2019.

\bibitem[Johnson et~al.(2017)Johnson, Hariharan, Van Der~Maaten, Fei-Fei, Lawrence~Zitnick, and Girshick]{johnson2017clevr}
Justin Johnson, Bharath Hariharan, Laurens Van Der~Maaten, Li Fei-Fei, C Lawrence~Zitnick, and Ross Girshick.
\newblock Clevr: A diagnostic dataset for compositional language and elementary visual reasoning.
\newblock In \emph{Proceedings of the IEEE conference on computer vision and pattern recognition}, pages 2901--2910, 2017.

\bibitem[Kembhavi et~al.(2016)Kembhavi, Salvato, Kolve, Seo, Hajishirzi, and Farhadi]{kembhavi2016diagram}
Aniruddha Kembhavi, Mike Salvato, Eric Kolve, Minjoon Seo, Hannaneh Hajishirzi, and Ali Farhadi.
\newblock A diagram is worth a dozen images.
\newblock In \emph{Computer Vision--ECCV 2016: 14th European Conference, Amsterdam, The Netherlands, October 11--14, 2016, Proceedings, Part IV 14}, pages 235--251. Springer, 2016.

\bibitem[Kim et~al.(2021)Kim, Yoo, Park, Kim, and Lee]{kim2021selfreg}
Daehee Kim, Youngjun Yoo, Seunghyun Park, Jinkyu Kim, and Jaekoo Lee.
\newblock Selfreg: Self-supervised contrastive regularization for domain generalization.
\newblock In \emph{Proceedings of the IEEE/CVF International Conference on Computer Vision}, pages 9619--9628, 2021.

\bibitem[Krueger et~al.(2021)Krueger, Caballero, Jacobsen, Zhang, Binas, Zhang, Le~Priol, and Courville]{krueger2021out}
David Krueger, Ethan Caballero, Joern-Henrik Jacobsen, Amy Zhang, Jonathan Binas, Dinghuai Zhang, Remi Le~Priol, and Aaron Courville.
\newblock Out-of-distribution generalization via risk extrapolation (rex).
\newblock In \emph{International conference on machine learning}, pages 5815--5826. PMLR, 2021.

\bibitem[Laboratory(2024)]{sharegpt4o}
Shanghai~AI Laboratory.
\newblock Sharegpt-4o, 2024.

\bibitem[Li et~al.(2018)Li, Pan, Wang, and Kot]{li2018domain}
Haoliang Li, Sinno~Jialin Pan, Shiqi Wang, and Alex~C Kot.
\newblock Domain generalization with adversarial feature learning.
\newblock In \emph{Proceedings of the IEEE conference on computer vision and pattern recognition}, pages 5400--5409, 2018.

\bibitem[Liang and Zou(2022)]{liang2022metashift}
Weixin Liang and James Zou.
\newblock Metashift: A dataset of datasets for evaluating contextual distribution shifts and training conflicts.
\newblock \emph{arXiv preprint arXiv:2202.06523}, 2022.

\bibitem[Lin et~al.(2014)Lin, Maire, Belongie, Hays, Perona, Ramanan, Doll{\'a}r, and Zitnick]{lin2014microsoft}
Tsung-Yi Lin, Michael Maire, Serge Belongie, James Hays, Pietro Perona, Deva Ramanan, Piotr Doll{\'a}r, and C~Lawrence Zitnick.
\newblock Microsoft coco: Common objects in context.
\newblock In \emph{Computer Vision--ECCV 2014: 13th European Conference, Zurich, Switzerland, September 6-12, 2014, Proceedings, Part V 13}, pages 740--755. Springer, 2014.

\bibitem[Liu et~al.(2023{\natexlab{a}})Liu, Lin, Li, Wang, Yacoob, and Wang]{liu2023aligning}
Fuxiao Liu, Kevin Lin, Linjie Li, Jianfeng Wang, Yaser Yacoob, and Lijuan Wang.
\newblock Aligning large multi-modal model with robust instruction tuning.
\newblock \emph{arXiv preprint arXiv:2306.14565}, 2023{\natexlab{a}}.

\bibitem[Liu et~al.(2023{\natexlab{b}})Liu, Wang, Yao, Chen, Song, Cho, Yacoob, and Yu]{liu2023mmc}
Fuxiao Liu, Xiaoyang Wang, Wenlin Yao, Jianshu Chen, Kaiqiang Song, Sangwoo Cho, Yaser Yacoob, and Dong Yu.
\newblock Mmc: Advancing multimodal chart understanding with large-scale instruction tuning.
\newblock \emph{arXiv preprint arXiv:2311.10774}, 2023{\natexlab{b}}.

\bibitem[Liu et~al.(2023{\natexlab{c}})Liu, Li, Wu, and Lee]{liu2023llava}
Haotian Liu, Chunyuan Li, Qingyang Wu, and Yong~Jae Lee.
\newblock Visual instruction tuning, 2023{\natexlab{c}}.

\bibitem[Liu et~al.(2024)Liu, Li, Li, and Lee]{liu2024improved}
Haotian Liu, Chunyuan Li, Yuheng Li, and Yong~Jae Lee.
\newblock Improved baselines with visual instruction tuning.
\newblock In \emph{Proceedings of the IEEE/CVF Conference on Computer Vision and Pattern Recognition}, pages 26296--26306, 2024.

\bibitem[Liu et~al.(2025)Liu, Duan, Zhang, Li, Zhang, Zhao, Yuan, Wang, He, Liu, et~al.]{liu2025mmbench}
Yuan Liu, Haodong Duan, Yuanhan Zhang, Bo Li, Songyang Zhang, Wangbo Zhao, Yike Yuan, Jiaqi Wang, Conghui He, Ziwei Liu, et~al.
\newblock Mmbench: Is your multi-modal model an all-around player?
\newblock In \emph{European Conference on Computer Vision}, pages 216--233. Springer, 2025.

\bibitem[Liu et~al.(2015)Liu, Luo, Wang, and Tang]{liu2015deep}
Ziwei Liu, Ping Luo, Xiaogang Wang, and Xiaoou Tang.
\newblock Deep learning face attributes in the wild.
\newblock In \emph{Proceedings of the IEEE international conference on computer vision}, pages 3730--3738, 2015.

\bibitem[Lu et~al.(2022)Lu, Mishra, Xia, Qiu, Chang, Zhu, Tafjord, Clark, and Kalyan]{lu2022learn}
Pan Lu, Swaroop Mishra, Tanglin Xia, Liang Qiu, Kai-Wei Chang, Song-Chun Zhu, Oyvind Tafjord, Peter Clark, and Ashwin Kalyan.
\newblock Learn to explain: Multimodal reasoning via thought chains for science question answering.
\newblock \emph{Advances in Neural Information Processing Systems}, 35:\penalty0 2507--2521, 2022.

\bibitem[Lu et~al.(2024)Lu, Bansal, Xia, Liu, Li, Hajishirzi, Cheng, Chang, Galley, and Gao]{lu2024mathvistaevaluatingmathematicalreasoning}
Pan Lu, Hritik Bansal, Tony Xia, Jiacheng Liu, Chunyuan Li, Hannaneh Hajishirzi, Hao Cheng, Kai-Wei Chang, Michel Galley, and Jianfeng Gao.
\newblock Mathvista: Evaluating mathematical reasoning of foundation models in visual contexts, 2024.

\bibitem[Lynch et~al.(2023)Lynch, Dovonon, Kaddour, and Silva]{lynch2023spawrious}
Aengus Lynch, Gb{\`e}tondji~JS Dovonon, Jean Kaddour, and Ricardo Silva.
\newblock Spawrious: A benchmark for fine control of spurious correlation biases.
\newblock \emph{arXiv preprint arXiv:2303.05470}, 2023.

\bibitem[Mishra et~al.(2019)Mishra, Shekhar, Singh, and Chakraborty]{mishra2019ocr}
Anand Mishra, Shashank Shekhar, Ajeet~Kumar Singh, and Anirban Chakraborty.
\newblock Ocr-vqa: Visual question answering by reading text in images.
\newblock In \emph{2019 international conference on document analysis and recognition (ICDAR)}, pages 947--952. IEEE, 2019.

\bibitem[Qiao and Low(2024)]{qiao2024understanding}
Rui Qiao and Bryan Kian~Hsiang Low.
\newblock Understanding domain generalization: A noise robustness perspective.
\newblock \emph{arXiv preprint arXiv:2401.14846}, 2024.

\bibitem[Radford et~al.(2021)Radford, Kim, Hallacy, Ramesh, Goh, Agarwal, Sastry, Askell, Mishkin, Clark, Krueger, and Sutskever]{radford2021learningtransferablevisualmodels}
Alec Radford, Jong~Wook Kim, Chris Hallacy, Aditya Ramesh, Gabriel Goh, Sandhini Agarwal, Girish Sastry, Amanda Askell, Pamela Mishkin, Jack Clark, Gretchen Krueger, and Ilya Sutskever.
\newblock Learning transferable visual models from natural language supervision, 2021.

\bibitem[Sagawa et~al.(2019)Sagawa, Koh, Hashimoto, and Liang]{sagawa2019distributionally}
Shiori Sagawa, Pang~Wei Koh, Tatsunori~B Hashimoto, and Percy Liang.
\newblock Distributionally robust neural networks for group shifts: On the importance of regularization for worst-case generalization.
\newblock \emph{arXiv preprint arXiv:1911.08731}, 2019.

\bibitem[Singh et~al.(2019)Singh, Natarajan, Shah, Jiang, Chen, Batra, Parikh, and Rohrbach]{singh2019towards}
Amanpreet Singh, Vivek Natarajan, Meet Shah, Yu Jiang, Xinlei Chen, Dhruv Batra, Devi Parikh, and Marcus Rohrbach.
\newblock Towards vqa models that can read.
\newblock In \emph{Proceedings of the IEEE/CVF conference on computer vision and pattern recognition}, pages 8317--8326, 2019.

\bibitem[Sun and Saenko(2016)]{sun2016deep}
Baochen Sun and Kate Saenko.
\newblock Deep coral: Correlation alignment for deep domain adaptation.
\newblock In \emph{Computer Vision--ECCV 2016 Workshops: Amsterdam, The Netherlands, October 8-10 and 15-16, 2016, Proceedings, Part III 14}, pages 443--450. Springer, 2016.

\bibitem[Vapnik(1991)]{vapnik1991principles}
Vladimir Vapnik.
\newblock Principles of risk minimization for learning theory.
\newblock \emph{Advances in neural information processing systems}, 4, 1991.

\bibitem[Yang et~al.(2023)Yang, Zhang, Katabi, and Ghassemi]{yang2023changehardcloserlook}
Yuzhe Yang, Haoran Zhang, Dina Katabi, and Marzyeh Ghassemi.
\newblock Change is hard: A closer look at subpopulation shift, 2023.

\bibitem[Yu et~al.(2023)Yu, Yang, Li, Wang, Lin, Liu, Wang, and Wang]{yu2023mm}
Weihao Yu, Zhengyuan Yang, Linjie Li, Jianfeng Wang, Kevin Lin, Zicheng Liu, Xinchao Wang, and Lijuan Wang.
\newblock Mm-vet: Evaluating large multimodal models for integrated capabilities.
\newblock \emph{arXiv preprint arXiv:2308.02490}, 2023.

\bibitem[Zhang et~al.(2023)Zhang, He, Xu, Yu, Shen, and Cui]{zhang2023nico++}
Xingxuan Zhang, Yue He, Renzhe Xu, Han Yu, Zheyan Shen, and Peng Cui.
\newblock Nico++: Towards better benchmarking for domain generalization.
\newblock In \emph{Proceedings of the IEEE/CVF conference on computer vision and pattern recognition}, pages 16036--16047, 2023.

\end{thebibliography}
}


\clearpage
\setcounter{page}{1}
\maketitlesupplementary

\section{Domain Generalization Results on Resnet18}
\label{sec:resnet}

We provide domain generalization results on \coin~using Resnet18 in Tab.~\ref{tab:resnet_dg},~\ref{tab:resnet_worstgroup_self} and ~\ref{tab:resnet_worstgroup_cross}. Different from Sec.~\ref{sec:DG} testing methods in a linear probing setting, we benchmark using Resnet18.

\paragraph{Experimental Details.} Similar to Sec.~\ref{sec:DG}, we follow the implementation of DomainBed \cite{gulrajani2020searchlostdomaingeneralization}. On CIFAR10, we train Resnet18 from scratch, and on Tiny-Imagenet, we use Resnet18 with weights pre-trained on ImageNet. \footnote{To ensure fair comparison, we adopt the DomainBed implementations, and we verified that on Resnet18, regular training recipe can reach 95\% of accuracy.} 

\paragraph{Experiment Configuration}
For our experiments, we adopted specific hyperparameter settings optimized for the datasets used. Below is a summary of the configurations:

\begin{itemize}
    \item \textbf{Learning Rate:} $1 \times 10^{-3}$
    \item \textbf{Batch Size:} 128
    \item \textbf{Weight Decay:} $1 \times 10^{-4}$
    \item \textbf{Learning Rate Schedulers:}
    \begin{itemize}
        \item \textbf{CIFAR-10}: Cosine Annealing LR Scheduler with a step size of 20,000, and minimum learning rate of 0
        \item \textbf{TinyImageNet}: Step LR Scheduler with a step size of 20,000, and gamma of 0.1
    \end{itemize}
    \item \textbf{Training Steps:} 60,000 steps in total
    \item \textbf{Validation Split:} 5\% for holdout fraction
\end{itemize}

\paragraph{GroupDRO is the best on generalization.} GroupDRO is the best overall (Tab.~\ref{tab:resnet_dg}) and shows noticeable superiority over other algorithms. 

\paragraph{ERM is the best with subpopulation shift, followed by GroupDRO.} Tab.~\ref{tab:resnet_worstgroup_self} presents the WGA when evaluated on the same set of subgroups as training. We observe that ERM and GroupDRO are on par with each other. \textit{Subgroup robust} methods like GroupDRO is a top performer, unlike in the linear probing setting. 
\paragraph{GroupDRO excels again at attribute generalization.} Tab.~\ref{tab:resnet_worstgroup_cross} shows that GroupDRO is consistently better than others in the attribute generalization setting. 

\begin{table}[h!]
\centering
\begin{adjustbox}{width=\linewidth}
\begin{tabular}{c|ccccccc|c}
\toprule
Trained$\times$Evaluated &   ERM &GroupDRO &ARM & CORAL & EQRM & SelfReg & VREx & Average\\ 
\midrule
C10FGD$\times$FGD &   85.00 &89.33 &84.67 & 85.71 & 85.71 & 84.76 & 88.05 & 86.18 \\
C10BGD$\times$BGD &   80.00 &81.33 &80.43 & 79.67 & 84.67 & 82.80 & 82.26 & 81.59\\
C10ENV$\times$ENV &   81.20 &79.66 &80.03 & 79.40 & 81.97 & 81.62 & 81.66 & 80.79\\
TIFGD$\times$FGD &   59.00 &58.17 &57.26 & 54.80 & 56.78 & 53.36 & 54.42 & 56.26\\
TIBGD$\times$BGD &   63.06 &62.41 &62.90 & 55.66 & 59.84 & 55.18 & 54.30 & 59.05\\
TIENV$\times$ENV &   62.16 &59.39 &62.62 & 56.62 & 60.68 & 55.82 & 57.09 & 59.20\\
\midrule
Average &   \textbf{71.74} & \underline{71.72}  &71.32 & 68.64 & 71.61 & 68.92 & 69.63 & 70.51\\\bottomrule
\end{tabular}
\end{adjustbox}
\caption{Worst group accuracy (\%) when using the same training and evaluation subgroups. ERM and GroupDRO are on par with each other.}
\label{tab:resnet_worstgroup_self}
\end{table}
\begin{table}[h!]
\centering
\begin{adjustbox}{width=\linewidth}
\begin{tabular}{lc|ccccccc|c}
\toprule
 \multicolumn{2}{c|}{Trained $\times$ Evaluated}&   ERM &GroupDRO &ARM & CORAL & EQRM & SelfReg & VREx & Average\\ 
\midrule
\multirow{5}{*}{FGD} &C10FGD$\times$BGD &   81.29 & 82.67 &79.33 & 79.67 & 82.67 & 80.00 & 81.18 & 80.97\\
 &C10FGD$\times$ENV &   82.96 &85.70 &81.94 & 82.88 & 85.19 & 82.99 & 83.40 & 83.58\\
 &TIFGD$\times$BGD &   58.71 &59.68 &58.87 & 54.22 & 55.66 & 50.36 & 53.23 & 55.82\\
 &TIFGD$\times$ENV &   60.84 &61.44 &60.60 & 55.16 & 57.70 & 54.82 & 55.45 & 58.00\\
\cmidrule{2-10}
 &Average &   67.98 &\textbf{72.37} &70.31 & 70.95 &  70.19 & 67.04 & 68.32 & 69.59 \\
\midrule
\multirow{5}{*}{BGD} &C10BGD$\times$FGD &   85.33 &86.67 &85.71 & 87.47 & 86.67 & 87.33 & 85.71 & 86.41\\
 &C10BGD$\times$ENV &   82.59 &86.21 &82.51 & 83.76 & 85.07 & 83.79 & 85.04 & 84.14\\
 &TIBGD$\times$FGD &   59.00 &60.93 &59.18 & 53.13 & 57.84 & 53.36 & 54.22 & 56.81 \\
 &TIBGD$\times$ENV &   62.21 &62.99 &61.71 & 54.89 & 59.37 & 55.96 & 56.03 & 59.02\\
\cmidrule{2-10}
 &Average &   72.20 &\textbf{74.20}  &72.28 & 69.81 & 72.24 &  70.11  & 70.25 & 71.60 \\
\midrule
\multirow{5}{*}{ENV} &C10ENV$\times$FGD &   82.67 &80.95 &81.00 & 77.14 & 82.18 & 82.38 & 85.42 & 81.68\\
 &C10ENV$\times$BGD &   77.00 &80.00 &75.00 & 75.33 & 79.67 & 79.67 & 80.38 & 78.15\\
 &TIENV$\times$FGD &   59.51 &60.37 &59.81 & 53.61 & 59.48 & 51.54 & 55.19 & 57.07\\
 &TIENV$\times$BGD &   60.16 &59.52 &62.89 & 55.48 & 59.35 & 54.83 & 53.49 & 57.96\\
\cmidrule{2-10}
 &Average &   69.84 &\textbf{70.21} &69.68 & 65.39 & 70.17 & 67.11 & 68.62 & 68.72\\\bottomrule
\end{tabular}
\end{adjustbox}
\caption{Worst group performance (\%) in attribute generalization (AG) setting. GroupDRO is the most effective in this setup.}
\label{tab:resnet_worstgroup_cross}
\end{table}
\begin{table*}[th]
\captionsetup{belowskip=-10pt}
    \centering
    \begin{adjustbox}{width=\linewidth}
\begin{tabular}{lcccccccccc|c}
\toprule
Method  & C10FGD  & C10FGDmerge  & C10BGD  & C10BGDmerge  & C10ENV  & TIFGD  & TIFGDmerge  & TIBGD  & TIBGDmerge  & TIENV  & Average\\
\midrule
 ERM  & 89.39(0.27)  & 89.94(0.26)  & 89.48(0.22)  & 90.31(0.19)  & 87.67(0.29)  & 66.38(0.32)  & 67.04(0.31)  & 67.45(0.32)  & 67.64(0.43)  & 66.98(0.41) &\underline{78.23}(0.30) \\
 GroupDRO  & 91.54(0.29)  & 91.65(0.13)  & 91.24(0.27)  & 91.31(0.30)  & 88.90(0.27)  & 67.13(0.32)  & 67.10(0.45)  & 68.18(0.37)  & 68.21(0.66)  & 67.42(0.28) &\textbf{79.27}(0.33) \\
 ARM  &  88.83(0.39)  &  89.23(0.15)  &  89.06(0.24)  &  89.81(0.30)  &  87.01(0.24)  &  66.23(0.48)  &  66.53(0.38)  &  67.29(0.74)  &  67.54(0.51)  &  67.09(0.36) & 77.86(0.38) \\
 CORAL  &  89.95(0.29)  &  90.56(0.24)  &  89.60(0.24)  &  90.64(0.08)  &  85.46(0.47)  &  60.56(4.99)  &  62.86(4.13)  &  59.51(6.41)  &  62.86(4.89)  &  59.43(6.22) & 75.14(2.80) \\
 EQRM  &  90.39(0.56)  &  90.70(0.33)  &  90.11(0.56)  &  90.78(0.32)  &  88.34(0.42)  &  63.09(2.19)  &  64.62(2.11)  &  64.59(2.01)  &  65.43(2.25)  &  66.08(2.27) & 77.41(1.30) \\
 SelfReg  &  89.86(0.25)  &  90.26(0.20)  &  89.96(0.13)  &  90.49(0.14)  &  88.19(0.10)  &  59.61(4.05)  &  61.30(2.73)  &  60.56(3.36)  &  62.72(2.24)  &  58.38(4.25) & 75.13(1.75) \\
 VREx &  90.38(0.75) &  90.99(0.66) &  90.57(1.07) &  90.83(0.51) &  89.25(0.84) &  61.22(3.53) &  61.27(3.84) &  61.64(3.60) &  62.35(3.47) &  62.45(2.62) & 76.10(2.09)\\
 \midrule
 Average & 90.05(0.40) & 90.48(0.28) & 90.00(0.39) & 90.60(0.26) & 87.83(0.38) & 63.46(2.27) & 64.39(1.99) & 64.17(2.40) & 65.25(2.06) & 63.98(2.34) & 77.02(1.28)\\
\bottomrule
\end{tabular}
\end{adjustbox}
\caption{Domain generalization algorithms results (\%, std) on \coin~(Resnet18). We observe that GroupDRO is the overall best.}
\label{tab:resnet_dg}
\end{table*}

\section{Evaluation Cost of GPT-4o}
The evaluation of \coin-Bench on GPT-4o costs \$560.
\section{Dataset Release}
We will release \coin~annotations upon the acceptance of this paper to facilitate research in VLM SFT and domain generalization.

%
%
\end{document}